\documentclass[preprint,12pt]{elsarticle}

\usepackage{epsfig}

\usepackage{amssymb}
\usepackage{amsmath}
\usepackage{amsthm}
\usepackage{caption}
\usepackage{subcaption}
\usepackage{mathabx}
\usepackage{color}

\newcommand{\ignore}[1]{}

\usepackage{lipsum}
\makeatletter
\def\ps@pprintTitle{%
 \let\@oddhead\@empty
 \let\@evenhead\@empty
 \def\@oddfoot{}%
 \let \@oddfoot}
\makeatother
\begin{document}

\begin{frontmatter}

%% Title, authors and addresses

%% use the tnoteref command within \title for footnotes;
%% use the tnotetext command for theassociated footnote;
%% use the fnref command within \author or \address for footnotes;
%% use the fntext command for theassociated footnote;
%% use the corref command within \author for corresponding author footnotes;
%% use the cortext command for theassociated footnote;
%% use the ead command for the email address,
%% and the form \ead[url] for the home page:
%% \title{Title\tnoteref{label1}}
%% \tnotetext[label1]{}
%% \author{Name\corref{cor1}\fnref{label2}}
%% \ead{email address}
%% \ead[url]{home page}
%% \fntext[label2]{}
%% \cortext[cor1]{}
%% \address{Address\fnref{label3}}
%% \fntext[label3]{}

\title{Design, Analysis and Application of A Volumetric Convolutional Neural Network}

%% use optional labels to link authors explicitly to addresses:
%% \author[label1,label2]{}
%% \address[label1]{}
%% \address[label2]{}

\author{Xiaqing Pan}
\author{Yueru Chen}
\author{C.-C. Jay Kuo}
\address{Ming-Hsieh Department of Electrical Engineering, University of Southern California, Los Angeles, CA 90089-2564, USA}

\begin{abstract}
%The design, analysis and application of a volumetric convolutional
%neural network (VCNN) are studied in this work. Although a large number
%of CNNs have been proposed in the literature, their design is empirical.
%In the design of the VCNN, we propose a feed-forward K-means clustering
%algorithm to determine the filter number and size at each convolutional
%layer systematically. For the analysis of the VCNN, we focus on the
%relationship between the filter weights (also known as anchor vectors)
%from the last fully connected (FC) layer to the output.  Typically, the
%output of the VCNN contains a couple of sets of confusing classes, and
%the cause of these confusion sets can be well explained by analyzing
%their anchor vector relationships. Furthermore, a hierarchical clustering
%method followed by a random forest classification method is proposed to
%boost the classification performance among confusing classes. For the
%application of the VCNN, we examine the 3D shape classification problem
%and conduct experiments on a popular dataset called the ModelNet40. The
%proposed VCNN offers the state-of-the-art performance among all
%volume-based CNN methods. 

The design, analysis and application of a volumetric convolutional neural network (VCNN) are studied in this work. Although many CNNs have been proposed in the literature, their design is empirical. In the design of the VCNN, we propose a feed-forward K-means clustering algorithm to determine the filter number and size at each convolutional layer systematically. For the analysis of the VCNN, the cause of confusing classes in the output of the VCNN is explained by analyzing the relationship between the filter weights (also known as anchor vectors) from the last fully-connected layer to the output. Furthermore, a hierarchical clustering method followed by a random forest classification method is proposed to boost the classification performance among confusing classes. For the application of the VCNN, we examine the 3D shape classification problem and conduct experiments on a popular ModelNet40 dataset. The proposed VCNN offers the state-of-the-art performance among all volume-based CNN methods. 
\end{abstract}

\begin{keyword}
%% keywords here, in the form: keyword \sep keyword
Convolutional neural network \sep 3D shape classification \sep
ModelNet40 shape dataset \sep unsupervised learning \sep anchor vector
%% PACS codes here, in the form: \PACS code \sep code

%% MSC codes here, in the form: \MSC code \sep code
%% or \MSC[2008] code \sep code (2000 is the default)

\end{keyword}

\end{frontmatter}

%% \linenumbers

%% main text
%% Major comments from CVPR and my answer
%%1. View-based methods outperform volume-based methods, why do we use volume-based methods
%%A: search "CVPRQ1", one modification in the introduction section.

%%2. VCNN's improvement is not significant. confusion set reclassification provides little improvements
%%A: The first comment is unfair because we already compared our method with VoxNet. We also justified that other volume-based method has a far more complex structure but we explored the potential of VoxNet in the experimental results section. I add a future work for defending the second comments "confusion set reclassification performance". Because our confusion set identification process is reasonable, we can expect to use more advanced methods to classify these confusion sets and get better results in the future. Search "CVPRQ2", one modification in the experimental results section.

%%3. more experiments on real-world datasets (3D CT images)
%%A: these experiments are hard to conduct. I will skip this comment.

%%4. Why not do grid search to choose filter parameters
%%A: search "CVPRQ4", two modifications in the introduction section.

%%5. Why not use deeper network
%%A: search "CVPRQ5", one modification in the related work section.

%%6. It is not clear why confusion/reclassification method is needed. end to end training is more preferable
%%A: search "CVPRQ6", one modification in the introduction section.

\section{Introduction} \label{sec.intro}

3D shape classification \cite{shilane2004princeton} is an important yet
challenging task arising in recent years. Larger repositories of 3D
models \cite{spaeth2016can} such as google sketchup and Yobi 3D have
been built for many applications. They include 3D printing, game design,
mechanical manufacture, medical analysis, and so on. To handle the
increasing size of repositories, an accurate 3D shape classification
system is in demand. 

Quite a few hand-craft features \cite{li2012shrec},
\cite{li2015comparison}, \cite{tangelder2008survey} were proposed to
solve the 3D shape classification problem before.  Interesting
properties of a 3D model are explored from different representations
such as views \cite{chaouch2007new}, \cite{chen2003visual},
\cite{ohbuchi2008salient}, volumetric data \cite{kazhdan2003rotation},
\cite{novotni2004shape}, \cite{sundar2003skeleton}, and mesh models
\cite{bronstein2011shape}, \cite{bronstein2010scale},
\cite{gal2007pose}, \cite{smeets2013meshsift}. However, these features
are not discriminative enough to overcome large intra-class variation
and strong inter-class similarity. In recent years, solutions based on
the convolutional neural network (CNN) \cite{krizhevsky2012imagenet},
\cite{lecun2015deep} have been developed for numerous computer vision
applications with significantly better performances.  As evidenced by
recent publications \cite{shapeNet}, \cite{savvashrec},
\cite{xie2015deepshape}, the CNN solutions also outperform traditional
methods relying on hand-craft features in the 3D shape classification
problem. 

A CNN method classifies 3D shapes using either view-based
\cite{shi2015deeppano}, \cite{su2015multi} or volume-based input data
\cite{maturana2015voxnet}, \cite{qi2016volumetric}, \cite{wu20153d}.  A
view-based CNN classifies 3D shapes by analyzing multiple rendered views
while a volume-based CNN conducts classification directly in the 3D
representation. Currently, the classification performance of the
view-based CNN is better than that of the volume-based CNN since the
resolution of the volumetric input data has to be much lower than that
of the view-based input data due to higher memory and computational
requirements of the volumetric input.  On the other hand, since
volume-based methods preserve the 3D structural information, it is
expected to have a greater potential in the long run. 

%%%%%%%%%%%%%%%%%%%%%%%%%%%%%%%%%%%%%%%%%%%%%%%%%%%%
\begin{figure}[!t]
\centering
\begin{subfigure}[b] {0.75\textwidth}
\centering
\includegraphics[width=\linewidth]{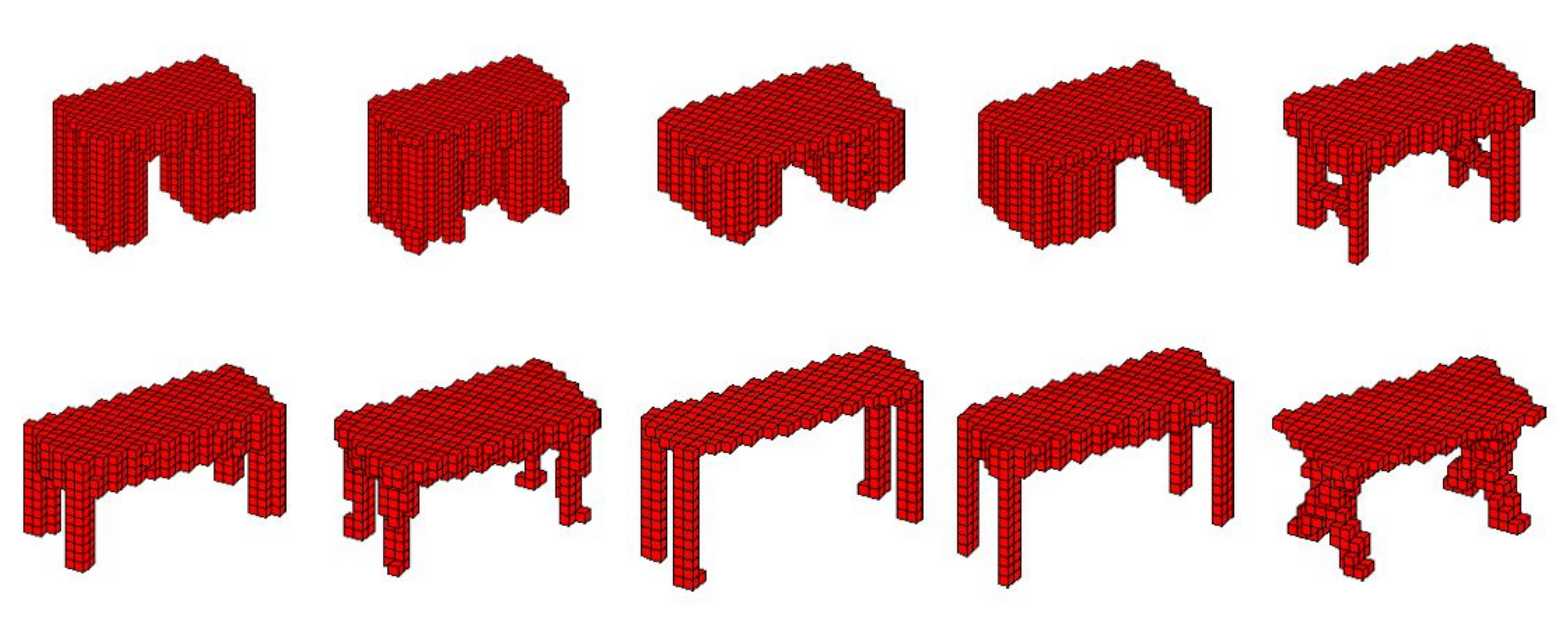}
\caption{Desk and table}
\end{subfigure}
\centering
\begin{subfigure}[b]{0.75\textwidth}
\includegraphics[width=\linewidth]{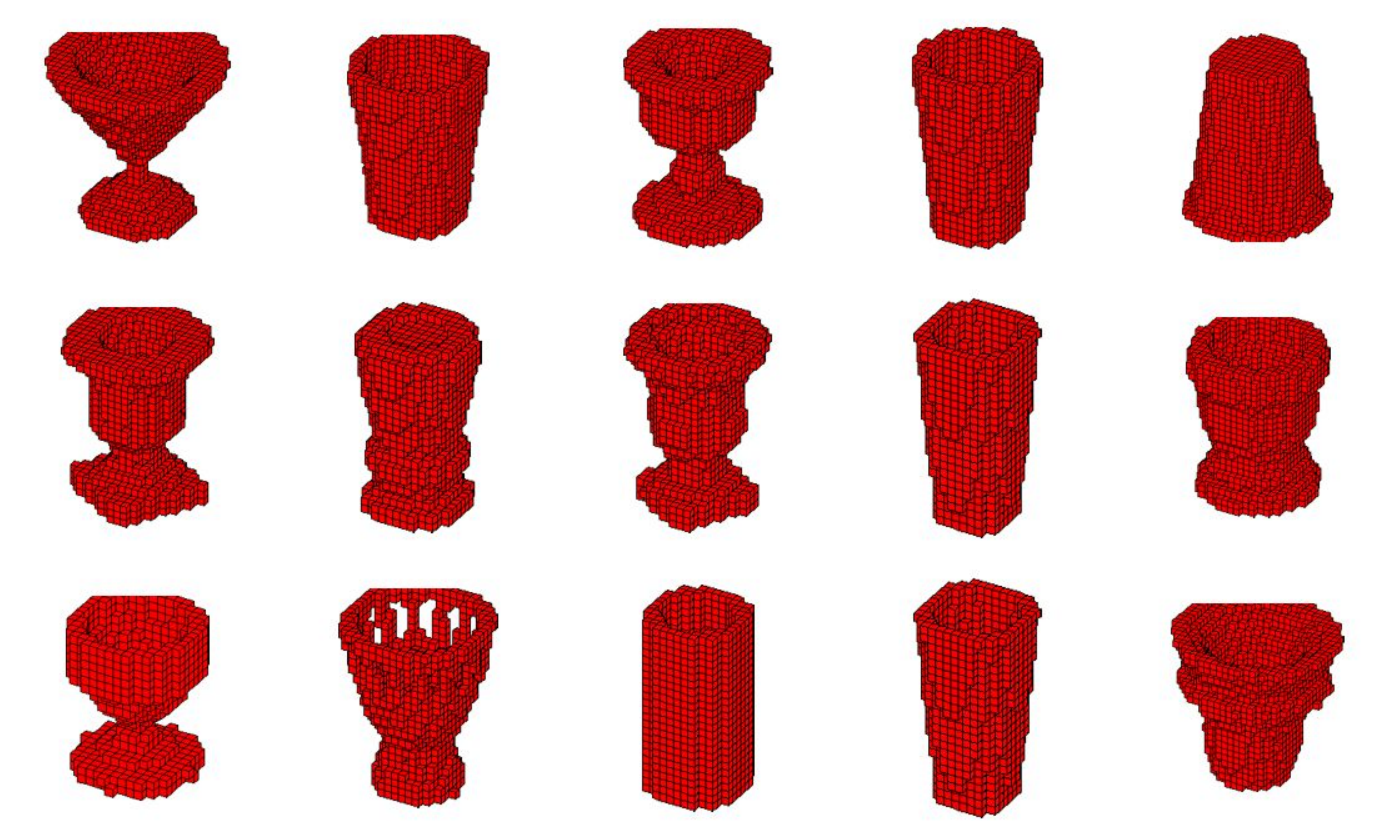}
\caption{Cup, flower pot and vase}
\end{subfigure}
\caption{Two sets of confusing classes: (a) desks (the first row)
and tables (the second row), and (b) cups (the first row), flower pots
(second row) and vases (third row). The confusion is attributed to the
similar global appearance of these 3D shapes.} \label{fig.confusion_cluster}
\end{figure}
%%%%%%%%%%%%%%%%%%%%%%%%%%%%%%%%%%%%%%%%%%%%%%%%%%%%

In this work, we target at shrinking the performance gap between
volume-based CNN methods and view-based CNN methods, and call the
resulting solution the ``Volumetric CNN" or VCNN in short.  Furthermore,
we attempt to conduct fundamental research on network design and
understanding in the following two aspects. 
\begin{enumerate}
\item The VCNN architecture is similar to that of the VoxNet
\cite{maturana2015voxnet}. In traditional CNN design, the network
parameters are chosen empirically with high computational complexity.
Here, we determine the VCNN network parameters with theoretic support.
Specifically, we propose a feed-forward K-means clustering algorithm to
identify the optimal filter number and the filter size systematically. 
\item Being similar to any other real world classification problem,
there exist sets of confusing classes in the 3D shape classification
problem \cite{dong2015looking}. We propose a method to determine whether
two shape classes could be confusing based on the network property
alone.  To enhance the classification performance among confusing
classes, we propose a hierarchical clustering method to split samples of
the same confusion set into multiple subsets. Then, samples in a subset
are reclassified using a random forest (RF) classifier. 
\end{enumerate}

To illustrate the second point above, we show two confusion sets in Fig.
\ref{fig.confusion_cluster}.  It will be argued that the filter weights
that connect the last fully connected (FC) layer and the output layer
carry very valuable information. It points to the centroid of feature
vectors of shapes in the same class, and is therefore called the "shape
anchor vector" (SAV) for the class. Two shape classes are confusing if
the angle of their SAVs is small and at least one of them has a large
inter-class variance (or diversity). 

The rest of this paper is organized as follows. Related work is reviewed
in Sec. \ref{sec.relateWork}. The proposed VCNN system is presented in
Sec. \ref{sec.VCNN}.  Experimental results are given to demonstrate the
superior performance of the VCNN method in Sec. \ref{sec.experiment}.
Finally, concluding remarks are given in Sec. \ref{sec.conclusion}. 

\section{Related Work}\label{sec.relateWork}

There are two main approaches proposed to classify 3D shape models: the
view-based approach and the volume-based approach. They are reviewed
below.  The view-based approach renders a 3D shape into multiple views
as the representation. Classifying a 3D shape becomes analyzing a bundle
of views collectively. The Multi-View Convolutional Neural Network
(MVCNN) \cite{su2015multi} method renders a 3D shape into 12 or 80
views. By adding a view-pooling layer in the VGG network model
\cite{Simonyan14c}, views of the input shape are merged before the fully
connected layers to identify salient regions and overcome the
rotational variance. The DeepPano \cite{shi2015deeppano} method
constructs panoramic views for a 3D shape. A row-wise max-pooling layer
is proposed to remove the shift variance. The MVCNN-Sphere method
\cite{qi2016volumetric} builds a resolution pyramid for each view using
sphere rendering based on the MVCNN structure. The classification
performance is improved by combining decisions from inputs of different
resolutions.  View-based methods can preserve the high resolution of 3D
shapes since they leverage the view projection and lower the complexity
from 3D to 2D. Furthermore, a view-based CNN can be fine-tuned from a
pretrained CNN that was trained by 2D images.  However, the view-based
CNN has two potential shortcomings. First, the surface of a 3D shape can
be affected by the shading effect.  Second, reconstructing the
relationship among views is difficult since the 3D information is lost
after the view-pooling process. 

The volume-based approach voxelizes a 3D mesh model for a 3D
representation. Several CNN networks have been proposed to classify the
3D shapes directly. Examples include the 3D ShapeNet \cite{wu20153d},
the VoxNet \cite{maturana2015voxnet} and the SubVolume supervision
method \cite{qi2016volumetric}. Although the network architectures of
volume-based methods are typically shallow (e.g., the VoxNet model
consists of two convolutional layers and one fully connected layer),
they strike a balance between the number of available training samples
and the network model complexity. Volume-based methods have two
drawbacks.  First, to control computational complexity, the resolution
of a 3D voxel model is much lower than that of its corresponding 2D
view-based representation.  As a result, high frequency components of
the original 3D mesh are sacrificed. Second, there are few pretrained
CNN models on 3D data, volume-based networks have to be trained from
scratch.  Although a volumetric representation preserves the 3D
structural information of an object, the performance of classifying 3D
mesh models directly is still lower than that of classifying the
corresponding view-based 2D models. 

\section{Proposed VCNN Method}\label{sec.VCNN}

\subsection{System Overview} \label{sec.overview}

\begin{figure}[t]
\centering
\includegraphics[width=0.70\linewidth]{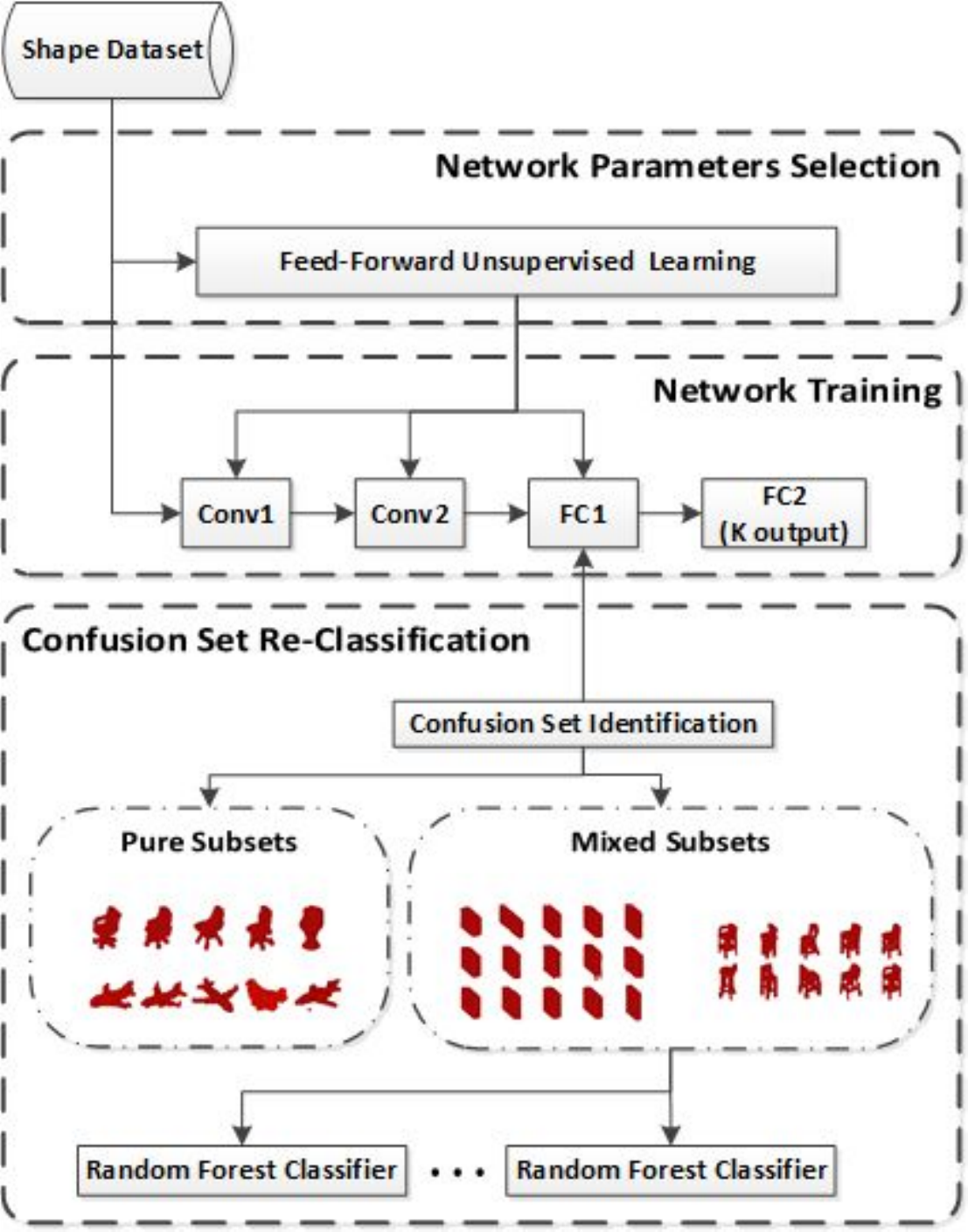}
\caption{The flow chart of the proposed system}\label{fig.flow_chart}
\end{figure}

An overview of the proposed VCNN solution is shown in Fig.
\ref{fig.flow_chart}. Before the supervised network training phase, we
first perform a feed-forward unsupervised clustering procedure to
determine the proper size and number of anchor vectors at each layer.
Then, in the training phase, the end-to-end backpropagation is
conducted, and all filter weights in the network are fine-tuned. The
feature vector of the last layer of the VCNN, called the the VCNN
feature, is acquired after the training process.  Afterwards, a
confusion matrix based on SAVs is used to identify confusing sets
categorized into pure sets and mixed sets. Each pure set contains a
single class of shapes. Each mixed set, which includes multiple classes
of samples, is split further into pure subsets and mixed subsets by a
proposed tree-structured hierarchical clustering algorithm. Each mixed
subset is trained by a random forest classifier.  In the testing phase,
a sample is assigned to a subset based on its VCNN feature. If the
sample is assigned to a pure subset, its label is immediately output.
Otherwise, it is assigned to a mixed subset and its label will be
determined by the random forest classifier associated with that subset. 

%%%%%%%%%%%%%%%%%%%%%%%%%%%%%%%%%%%%%%%%%%%%%%%%%%%%
%%%%%%%%%%%%%%%%%%%%%%%%%%%%%%%%%%%%%%%%%%%%%%%%%%%%

\subsection{Shape Anchor Vectors (SAVs)} \label{subsec.RECOS}

One key tool adopted in our work is the RECOS (Rectified-COrrelations on
a Sphere) model for CNNs as proposed in \cite{CCKuo}.  The weights of a
filter at intermediate layers are interpreted as a cluster centroid of
the corresponding inputs. Thus, these weights define an "anchor
vector".  The convolution and nonlinear activation operations at one
layer is viewed as projection onto a set of anchor vectors followed by
rectification in the RECOS model. The number of anchor filters is
related to the approximation capability of a certain layer. The more the
number of anchor vectors, the better the approximation capability yet
the higher the computational complexity. We will find a way to balance
approximation accuracy and computational complexity in Sec.
\ref{sec.design}. 

%%%%%%%%%%%%%%%%%%%%%%%%%%%%%%%%%%%%%%%%%%%%%%%%%%%%
\begin{figure}[t]
\centering
\includegraphics[width=0.6\linewidth]{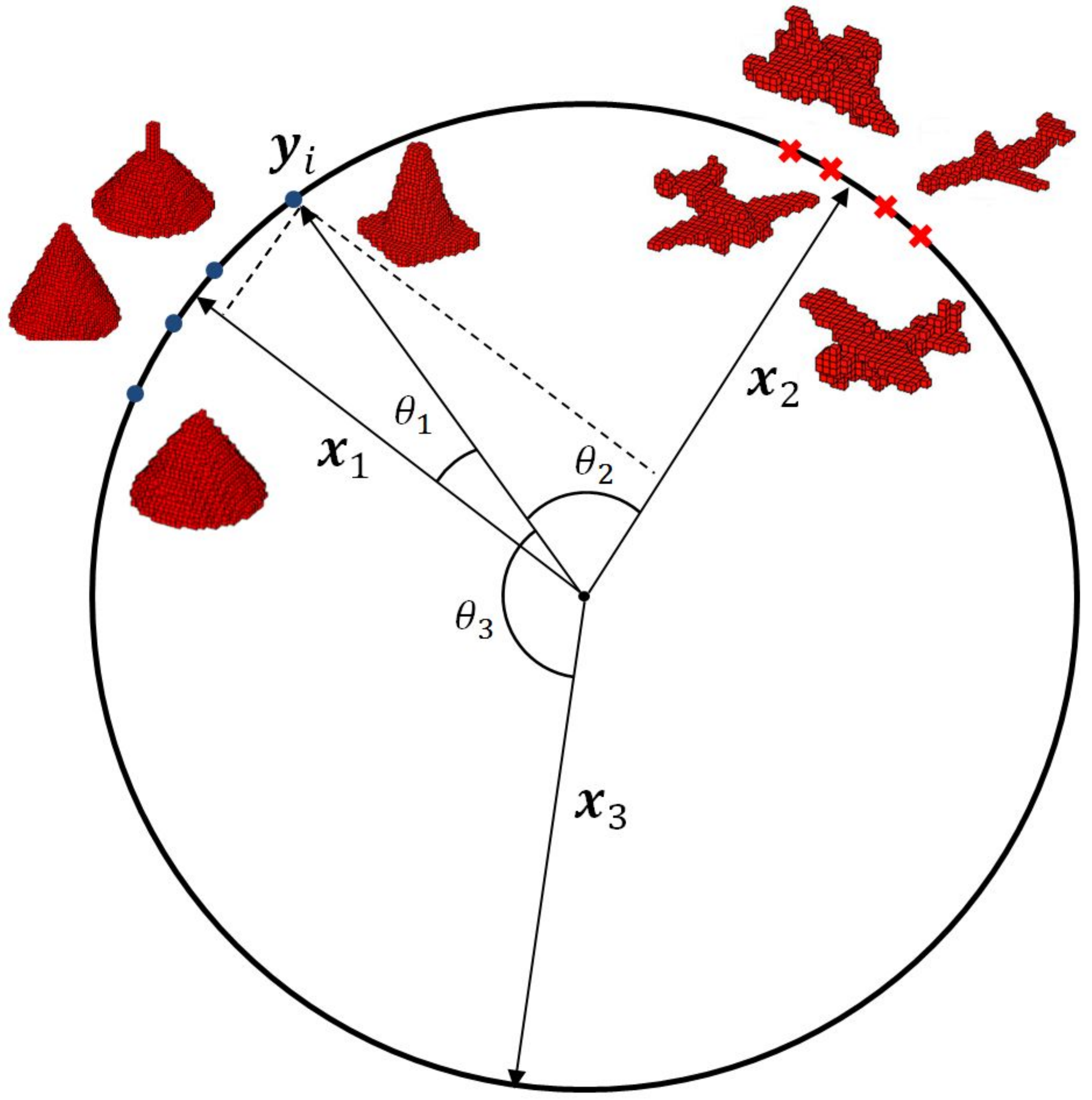}
\caption{Illustration of anchor vectors at the last stage of a CNN,
where each anchor vector points to a 3D shape class. They are called
the shape anchor vectors (SAVs). In this example, one SAV points to the
airplane class while another SAV points to the cone class.}\label{fig.circle}
\end{figure}
%%%%%%%%%%%%%%%%%%%%%%%%%%%%%%%%%%%%%%%%%%%%%%%%%%%%

The anchor vectors at the last stage of a CNN connect the last FC layer
to the output. They have a special physical meaning as shown in Fig.
\ref{fig.circle}, where each anchor vector points to a 3D shape class.
Thus, they are called the shape anchor vectors (SAVs). In this example,
one SAV points to the airplane class while another SAV points to the
cone class.  There are four blue dots and four red crosses along the
surface of a high-dimensional sphere. They represent the feature vectors
of 3D shape samples from the cone and the airplane classes,
respectively. The classification of each sample to a particular class is
determined by the shortest geodesic distance between the sample and the
tip of the SAV of the selected class (or, equivalently, the maximum
correlation between the sample vector and the SAV). Thus, sample $y_i$
is classified to the cone class as shown in this figure. 

The relationship between two SAVs and the sample distribution of a
particular class plays a critical role in determining whether two
classes are confusing or not. If the angle between two SAVs is small and
the samples of a particular class are distributed over a wider range, we
will get confusing classes. This will be elaborated in Sec.
\ref{sec.identifying}. 

\subsection{Network Parameters Selection} \label{sec.design}

The determination of network parameters such as the filter size and
number per layer is often conducted in an ad hoc manner in the current
literature. Here, we propose a systematic method to decide these
parameters based on a feed-forward unsupervised splitting approach. It
consists of two steps: 1) representative sample selection, and 2)
representative sample clustering. 

{\bf Problem Formulation.} We can formulate the network design problem
as follows. The input to the $j^{th}$ RECOS (or the $j$th convolutional
layer) has the following dimension:
$$
M^{j-1} \times N^{j-1} \times D^{j-1} \times K^{j-1}.
$$ 
where $M^{j-1} \times N^{j-1} \times D^{j-1}$ are spatial dimensions
while $K^{j-1}$ represents the spectral dimension (or the number of
anchor vectors) in the previous layer. Its output has the following
dimension:
$$
M^{j} \times N^{j} \times D^{j} \times K^{j}.
$$ 
Since the $2 \times 2 \times 2$ to $1 \times 1 \times 1$ maximum pooling
is adopted, we have $M^{j}= 0.5 M^{j-1}, N^{j}= 0.5 N^{j-1}, D^{j}= 0.5
D^{j-1}.$ Furthermore, the filter from the input to the output has a
dimension of
$$
m^{j} \times n^{j} \times d^{j} \times K^{j}.
$$
We set $m^j = n^j = d^j.$ This is because the input volumetric data is
of the cubic shape with the same resolution in all three spatial
dimension. Note $(M^{0}, N^{0}, D^{0}, K^{0})=(30,30,30,1)$ is the
dimension of the input 3D shape.  Given $(M^{j-1}, N^{j-1}, D^{j-1},
K^{j-1})$, the question is how to determine the filter size parameter,
$m^{j}$, and the filter number, $K^{j}$. 

{\bf Representative sample selection.} The dimension of the input vector
is $m^j \times n^j \times d^j \times K^{j-1} = (m^j)^3 \times K^{j-1}$.
Let $Y^{j-1}$ denotes the set of all training input samples.  However,
these input samples are not equally important. Some samples have less
discriminant power than others.  A good sample selection algorithm can
guide the network to learn a more meaningful way to partition the
feature space to select more discriminant SAVs for better decisions.

We first normalize all patterns $y^{j-1} \in Y^{j-1}$ to have the unit 
length. Then, the set $P^{j-1}$ of all normalized inputs can be written as
$$
P^{j-1} = \{ p^{j-1} \, | \, p^{j-1} = y^{j-1}/||y^{j-1}||, 
\, y^{j-1} \in Y^{j-1} \}.
$$
Next, we adopt the saliency principle in representative pattern
selection.  That is, we compute the variance of elements in $p^{j-1}$
and choose those of a larger variance value. If the variance of an input
is small, it has a flat element-wise distribution and, thus, has a low
discriminant power. 

We propose a two-stage screening process.In the first stage, a small
threshold $\epsilon$ is adopted to remove inputs with very small
variance values. That is, if the variance of an input is less than
$\epsilon$, it is removed from the candidate sample set.  In the second
stage, we select top $T\%$ samples with the largest variance values in
the filtered candidate sample set to keep all remaining candidates
sufficiently salient. 

After the above two-stage screening process, we have a smaller candidate
sample set that contain salient samples. If we want to reduce the size
of the candidate set furthermore to simplify the following computations,
a uniform random sampling strategy can be adopted.

{\bf Representative sample clustering.} Our objective is to find $K^j$
anchor vectors from the candidate set of representative samples obtained
from the above step. Here, we adopt an unsupervised clustering algorithm
such as the K-means algorithm to group them. The cluster centroids are
then set to the desired anchor vectors for the current RECOS model. 

To choose the optimal parameters, $m^j$ and $K^j$, we can maximize the
inter-cluster margin and minimize the intra-cluster variance by adopting
the Bayesian Information Criterion (BIC) function \cite{pelleg2000x}.
The approximation of the BIC function in the K-means algorithm can
be expressed as
\begin{align} \label{eq.BIC}
BIC(m^j, K^j, P^{j-1}) = -2\sum_{k_j = 1}^{K^j}{L_{k_j}} + 2K^jC^j\log{N},  
\end{align} 
where $C^j = (m^j)^3 \times K^{j-1}$ is the dimension of input vectors,
$N$ is the total number of elements in $P^{j-1}$, and $L$ is the
log-likelihood distance. Under the normal distribution model, we have
\begin{align} \label{eq.LogLikelihood} 
L_{k_j} = -\frac{N_{k_j}}{2}\sum_{c_j=1}^{C_j}\log(\delta_{c_j}^2 + 
\delta_{c_jk_j}^2),
\end{align} 
where $N_{k_j}$ is the number of samples in the $k_j$th cluster,
$\delta_{c_j}^2$ is the variance of the $c_j$th feature variable over
all samples, and $\delta_{c_jk_j}^2$ is the variance of the $c_j$th
feature variable in the $k_j$th cluster. By fixing parameter $C_j$, the
optimal value of $K_j$ is determined by detecting the valley point of
the BIC function. 

The relationship between two consecutive RECOS models is worthy
discussion. The problem of selecting optimal parameter pairs, $(m^j,
K^j)$, across multiple layers is actually inter-dependent.  Here, we
adopt a greedy algorithm to predict these parameters in a feed-forward
manner. The centroids after each K-means clustering serve as the anchor
vectors of each RECOS model and they can be used to generate input
samples for the next RECOS model. 

\subsection{Confusion Sets Identification and Re-Classification} \label{sec.identifying}

In principle, the network parameter selection algorithm presented in the
last subsection can be used to decide optimal network parameters for the
RECOS models in the leading layers.  As the network goes to the end, we
need to consider another factor. That is, the number of SAVs in the last
layer has to be equal to the number of classes due to the supervised
learning framework.  The discriminant power of the feature space before
the output layer can still be limited and, as a result, SAVs in the last
stage may not be able to separate 3D shapes of different classes
completely. Then, shapes from different classes can be mixed, leading to
confusion sets. In this subsection, we will develop a method to identify
these confusion sets. In particular, we would like to address the
following two questions: 1) how to split 3D shapes of the same class
into multiple sub-classes according to their VCNN features? and 2) what
is the relationship between SAVs and visually similar shapes? 

{\bf Generation of Sub-classes.} For the first question, we adopt a
tree-structured hierarchical clustering algorithm to split a class into
sub-classes, where the class can be either the ground-truth or the
predicted one. The algorithm initially splits the feature space into two
clusters according to their Euclidean distance by using the K-means
algorithm with $K=2$ \cite{hartigan1979algorithm}.  Then, for each
cluster, the variance is calculated to determine its tightness. A
cluster with a large variance is split furthermore. The termination of
the splitting process is decided by one of the following two criteria:
1) reaching a sufficiently small variance (say, below threshold
$\zeta$), and 2) reaching a sufficiently small size (say, below
threshold $\eta$). The number of sub-classes under each class can be
automatically determined by the two pre-selected threshold values,
$\zeta$ and $\eta$. 

%%%%%%%%%%%%%%%%%%%%%%%%%%%%%%%%%%%%%%%%%%%%%%%%%%%%
\begin{figure*}[t]
\centering
\begin{subfigure}[b] {0.4\textwidth}
\centering
\includegraphics[width=0.9\textwidth]{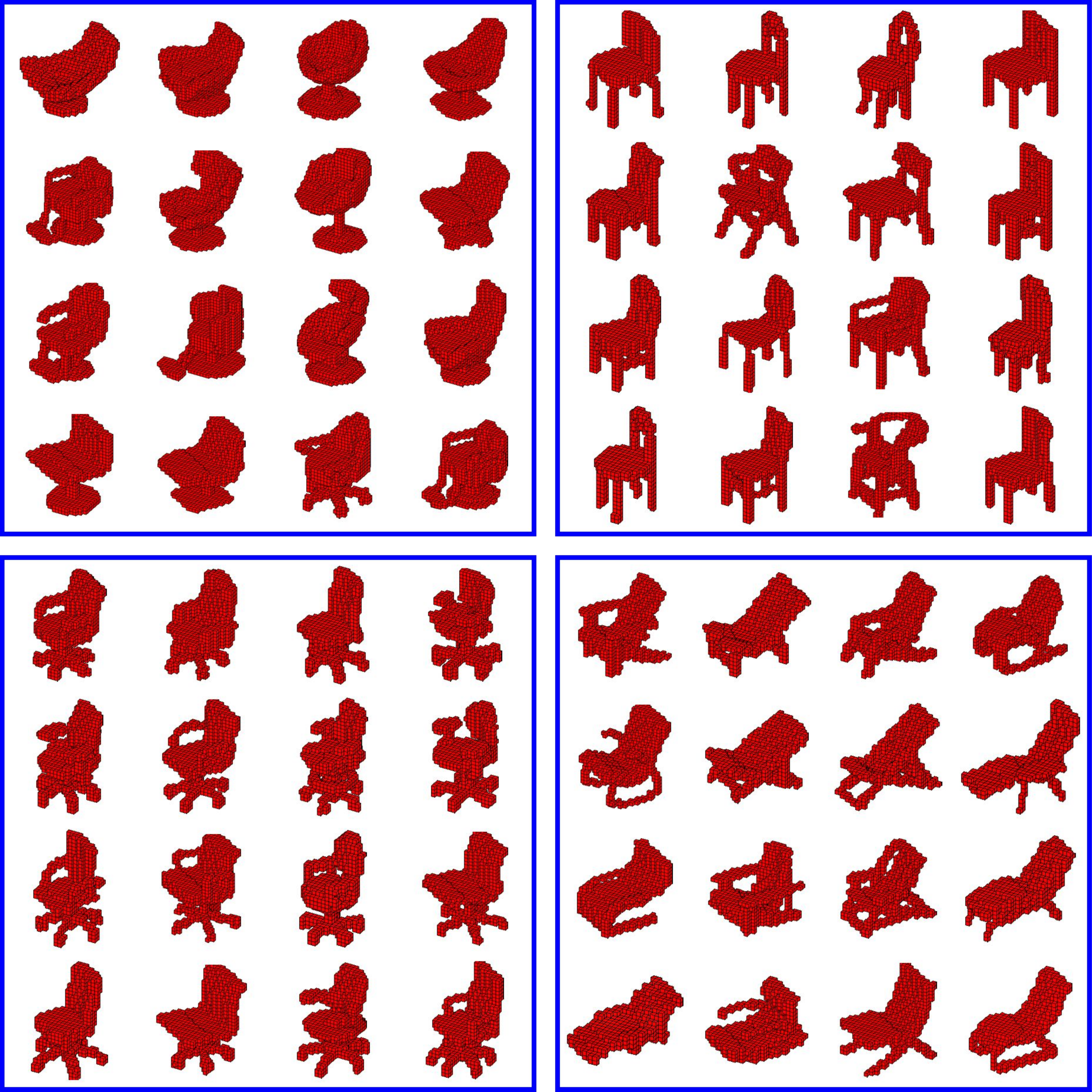}
\caption{Chair}
\end{subfigure}
\centering
\begin{subfigure}[b] {0.4\textwidth}
\centering
\includegraphics[width=0.9\textwidth]{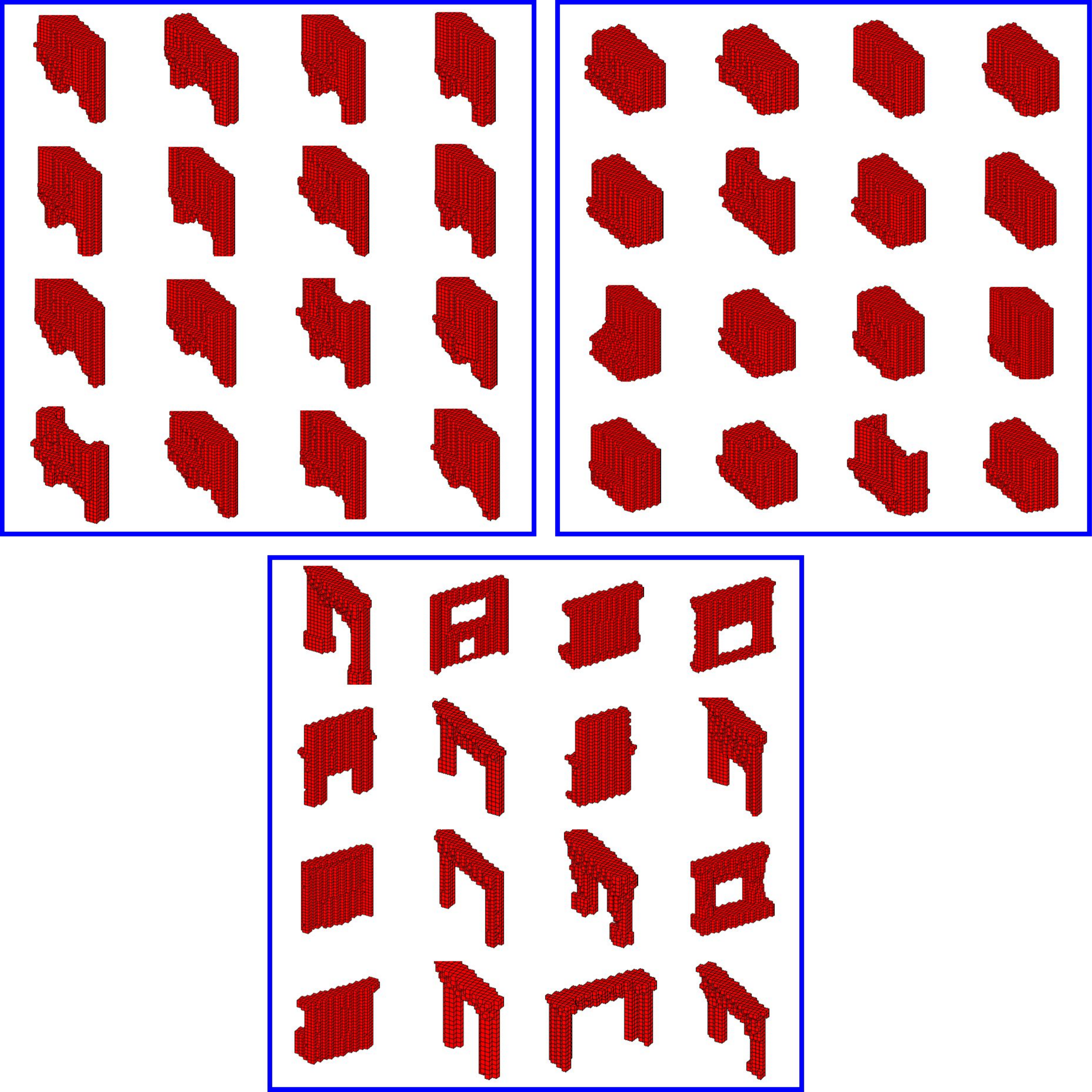}
\caption{Mantel}
\end{subfigure}
\centering
\begin{subfigure}[b] {0.4\textwidth}
\centering
\includegraphics[width=0.9\textwidth]{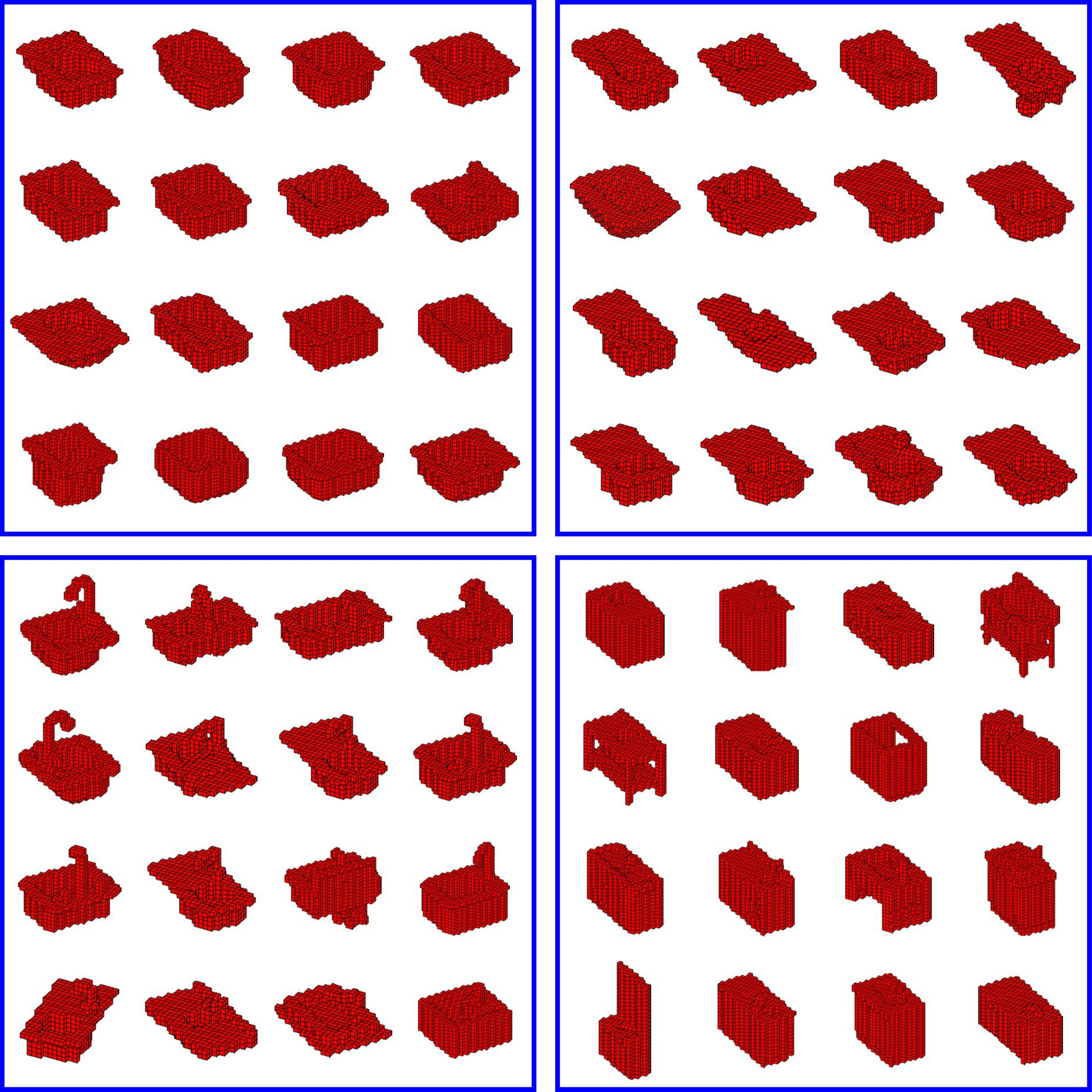}
\caption{Sink}
\end{subfigure}
\caption{Illustration of 3D shapes in sub-classes obtained from (a) the
Chair class, (b) the Mantel class and (c) the Sink class. We provide 16
representative shapes for each sub-class and encircle them with a blue
box.}\label{fig.subclass}
\end{figure*}
%%%%%%%%%%%%%%%%%%%%%%%%%%%%%%%%%%%%%%%%%%%%%%%%%%%%

We examine the power of the VCNN features and the tree-structure
hierarchical clustering algorithm using the ground-truth label.  Results
for three classes are shown in Fig. \ref{fig.subclass}. They are chair,
mantel and sink. We see that the VCNN features can generate sub-classes
that contain visually similar shapes automatically. For example, the
chair class can be divided into four sub-classes.  They are round
chairs, chairs with four feet, wheel chairs and long chairs.  The mantel
class can be further partitioned into three sub-classes. They are thin
mantels, thick mantels and hollow mantels. The sink class can be
clustered into four sub-classes; namely, deep sinks, flat sinks, sinks
with a hose and table-like sinks.  This shows the discriminant power of
the VCNN features, which are highly optimized by the backpropagation of
the CNN, and the power of the clustering algorithm in identifying
intra-class differences. 

{\bf Confusion Matrix Analysis and Confusion Set Identification.} After
running samples through the CNN, we will get predicted labels (instead
of ground-truth labels). Then, samples of globally similar appearance
but from different classes can be mixed in the VCNN feature space. This
explains the source of confusion. To resolve confusion among multiple
confusing classes, we adopt a merge-and-split strategy. That is, we
first merge several confusing classes into one confusion set and, then,
split the confusion set into multiple subsets as done before. 

We denote the directional confusion score of $i^{th}$ sample, $y_i$, to
the $k^{th}$ class, $c_k$, as $s(y_i, c_k)$. Theoretically, it is reciprocal to the projection distance between $y_i$ and the SAV of $c_k$. By normalizing the projection distances from $y_i$ to the SAVs of all classes by using the softmax function, the directional confusion score $s(y_i, c_k)$ is equivalent to the soft decision score.
The confusion factor (CF) between two classes $c_k$ and $c_l$ is determined
by the average of two directional confusion scores as
\begin{align} \label{eq.confMat}
CF(c_k, c_l) = \frac{1}{2N_k}\sum_{y_i \in c_k} s(y_i, c_l) + \frac{1}{2N_l}\sum_{y_j \in c_l} s(y_j, c_k),
\end{align}
where $N_k$ and $N_l$ are the numbers of samples in $c_k$ and $c_l$,
respectively. This CF value can be computed using training samples.

%%%%%%%%%%%%%%%%%%%%%%%%%%%%%%%%%%%%%%%%%%%%%%%%%%%%
\begin{figure}[t]
\centering
\includegraphics[width=0.9\linewidth]{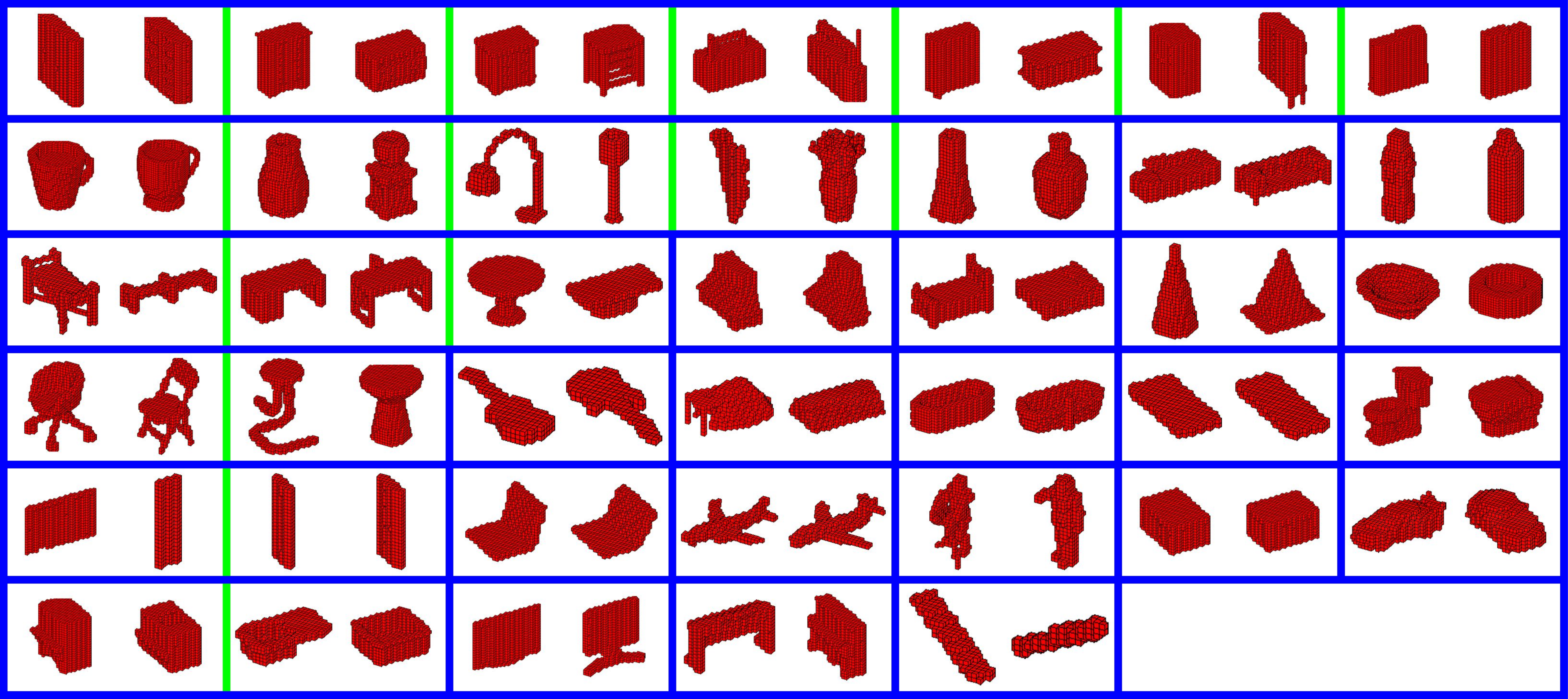}
\caption{A mixed or pure set is enclosed by a blue box.  Each mixed set contains multiple shape classes which are separated by
green vertical bars. Two representative 3D shapes are shown for each
class. Each row has one mixed set and several pure sets. The mixed set in the first row contains bookshelf, wardrobe, night
stand, radio, xbox, dresser and tv stand; that in the second row
contains cup, flower pot, lamp, plant and vase; that in the third row
contains bench, desk and table; that in the four row contains chair and
stool; that in the fifth row contains curtain and door; and that in the
six row contains mantel and sink.}\label{fig.confgroups}
\end{figure}
%%%%%%%%%%%%%%%%%%%%%%%%%%%%%%%%%%%%%%%%%%%%%%%%%%%%

Since the confusion matrix defines an affinity matrix, we adopt the
spectral clustering algorithm \cite{ng2002spectral} to cluster samples
into multiple confusion sets that have strong confusion
factors.  After the spectral clustering algorithm, we obtain either pure
or mixed sets. A pure set contains 3D shapes from the same class. Fig. \ref{fig.confgroups} shows the results of both mixed and pure sets obtained by the confusion matrix analysis. Each confusion set is
enclosed by a blue box. Inside a confusion set, each class is
represented by two instances. Two different classes are separated by a
green bar.  Some pure sets are generated because they are isolated from
other classes in the clustering process. It is worthwhile to emphasize that
each mixed set contains 3D shapes of similar appearance yet under
different class labels.  For example, the mixed set in the first row
contains seven classes: bookshelf, wardrobe, night stand, radio, xbox,
dresser and tv stand. All of them are cuboid like. 

In the testing phase, if a test sample is classified to a pure set,
the class label of that set is output as the desired answer.  A mixed
set contains 3D shapes from multiple classes, and further processing
is needed to separate them. This will be discussed below. 

%%%%%%%%%%%%%%%%%%%%%%%%%%%%%%%%%%%%%%%%%%%%%%%%%%%%
\begin{figure}[t]
\centering
\begin{subfigure}[b] {0.3\linewidth}
\centering
\includegraphics[width=0.95\linewidth]{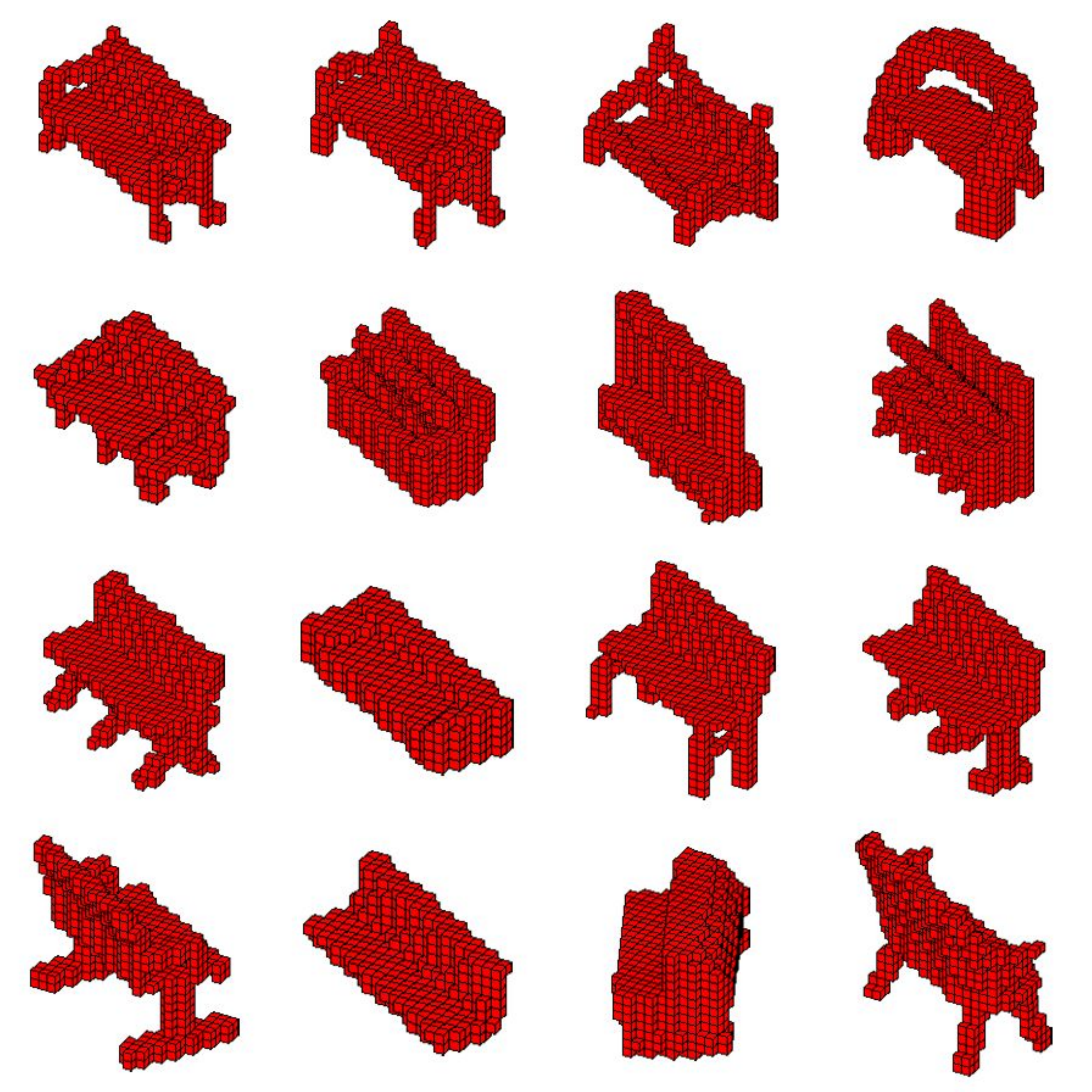}
\caption{ }
\end{subfigure}
\centering
\begin{subfigure}[b] {0.3\linewidth}
\centering
\includegraphics[width=0.95\linewidth]{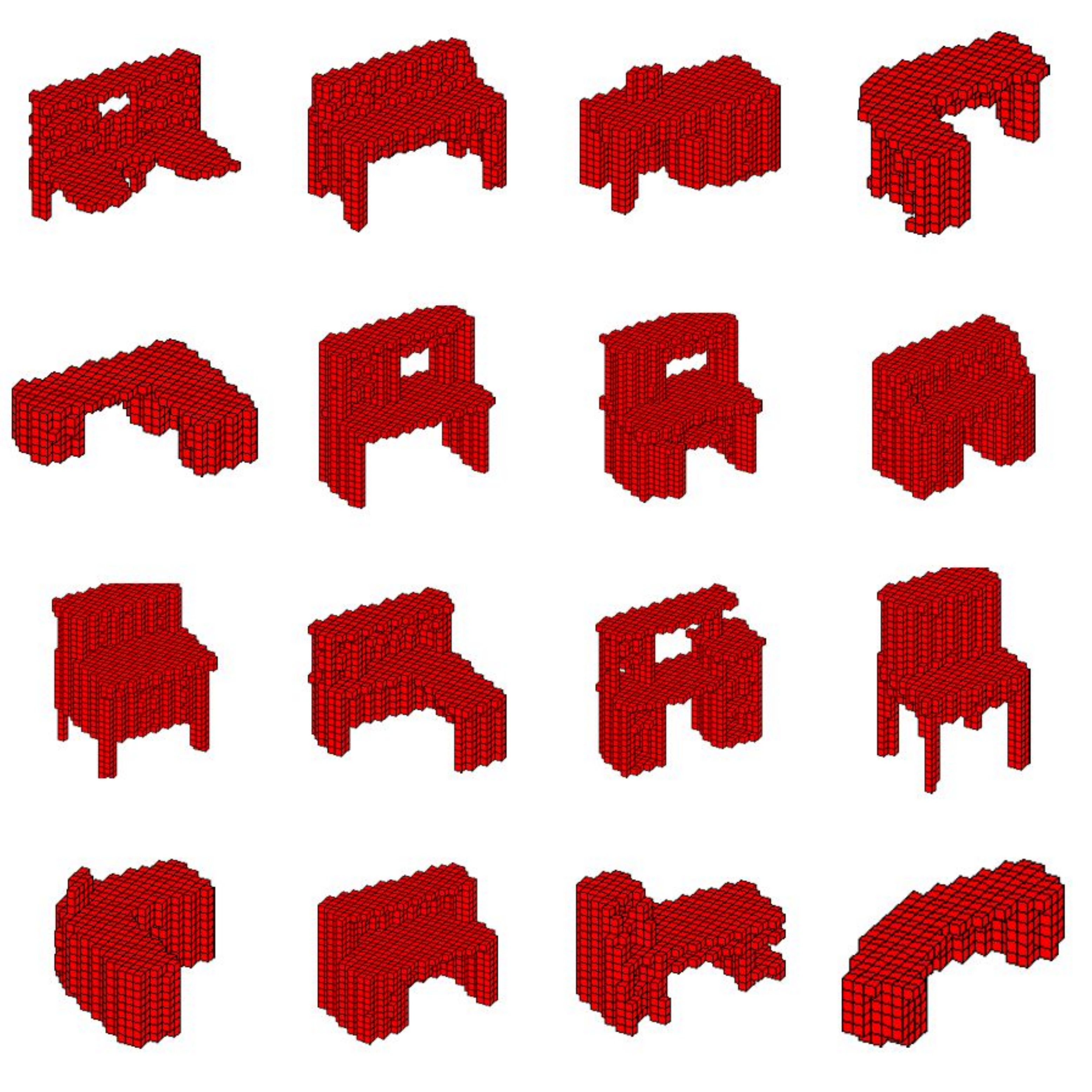}
\caption{ }
\end{subfigure}
\centering
\begin{subfigure}[b] {0.3\linewidth}
\centering
\includegraphics[width=0.95\linewidth]{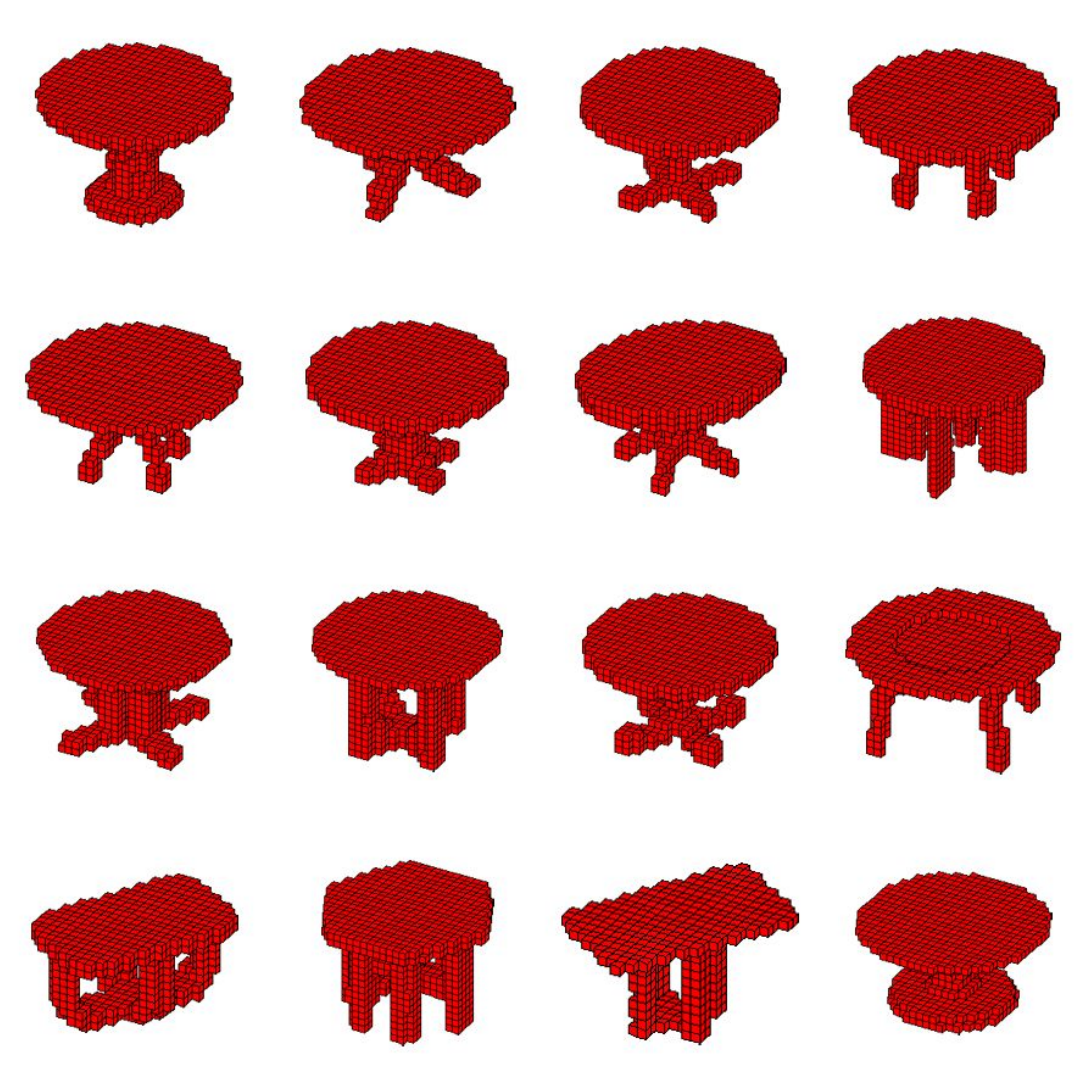}
\caption{ }
\end{subfigure}
\centering
\begin{subfigure}[b] {0.3\linewidth}
\centering
\includegraphics[width=0.95\linewidth]{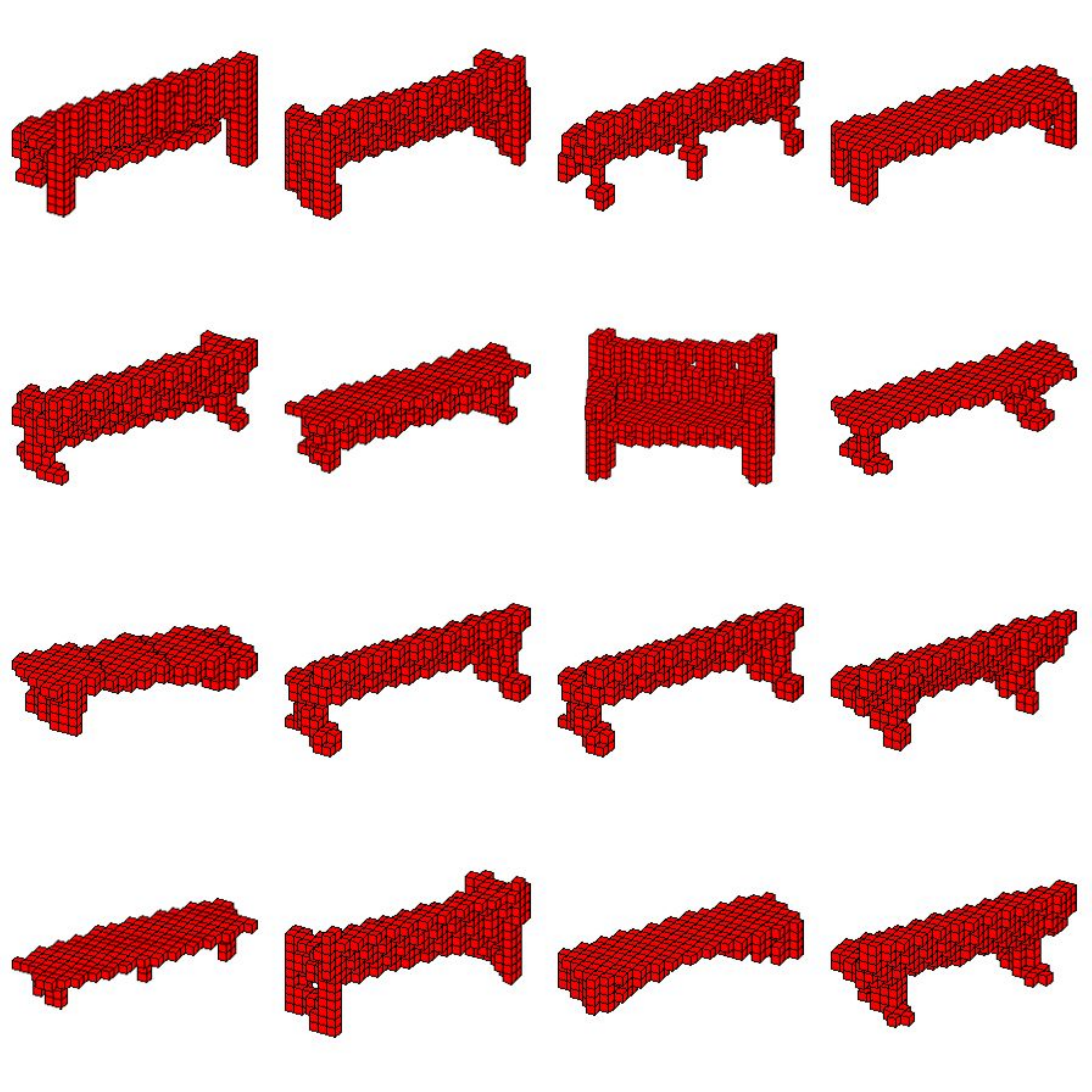}
\caption{ }
\end{subfigure}
\centering
\begin{subfigure}[b] {0.3\linewidth}
\centering
\includegraphics[width=0.95\linewidth]{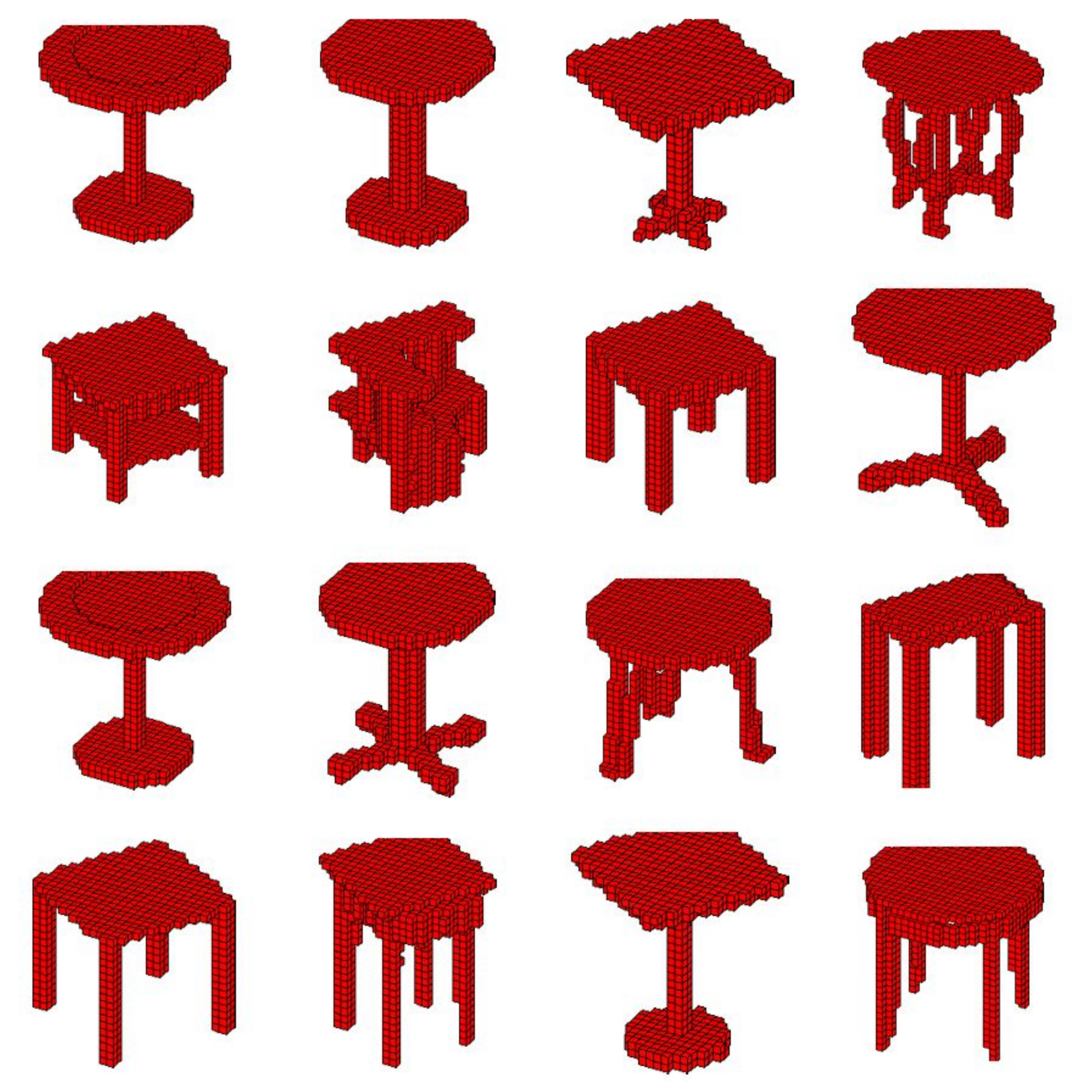}
\caption{ }
\end{subfigure}
\centering
\begin{subfigure}[b] {0.3\linewidth}
\centering
\includegraphics[width=0.95\linewidth]{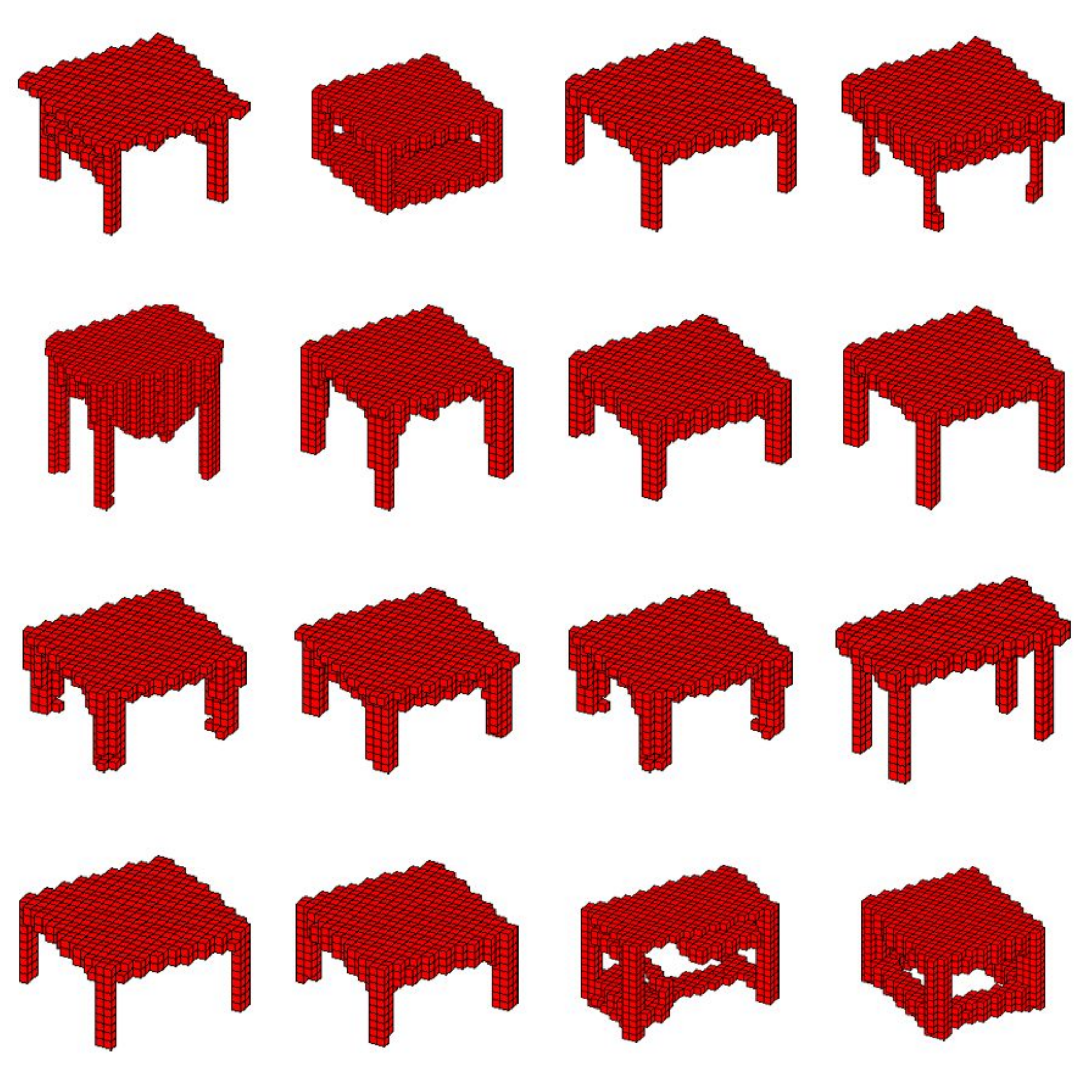}
\caption{ }
\end{subfigure}
\caption{The split of the confusion set of bench, desk and table yields
three pure subsets (a) bench, (b) desk and (c) table and three
mixed subsets (d) bench and table, (e) desk and table and (f)bench, desk 
and table. }\label{fig.confClustering1}
\end{figure}
%%%%%%%%%%%%%%%%%%%%%%%%%%%%%%%%%%%%%%%%%%%%%%%%%%%%
%%%%%%%%%%%%%%%%%%%%%%%%%%%%%%%%%%%%%%%%%%%%%%%%%%%%
\begin{figure}[t]
\centering
\begin{subfigure}[b] {0.3\linewidth}
\centering
\includegraphics[width=0.95\linewidth]{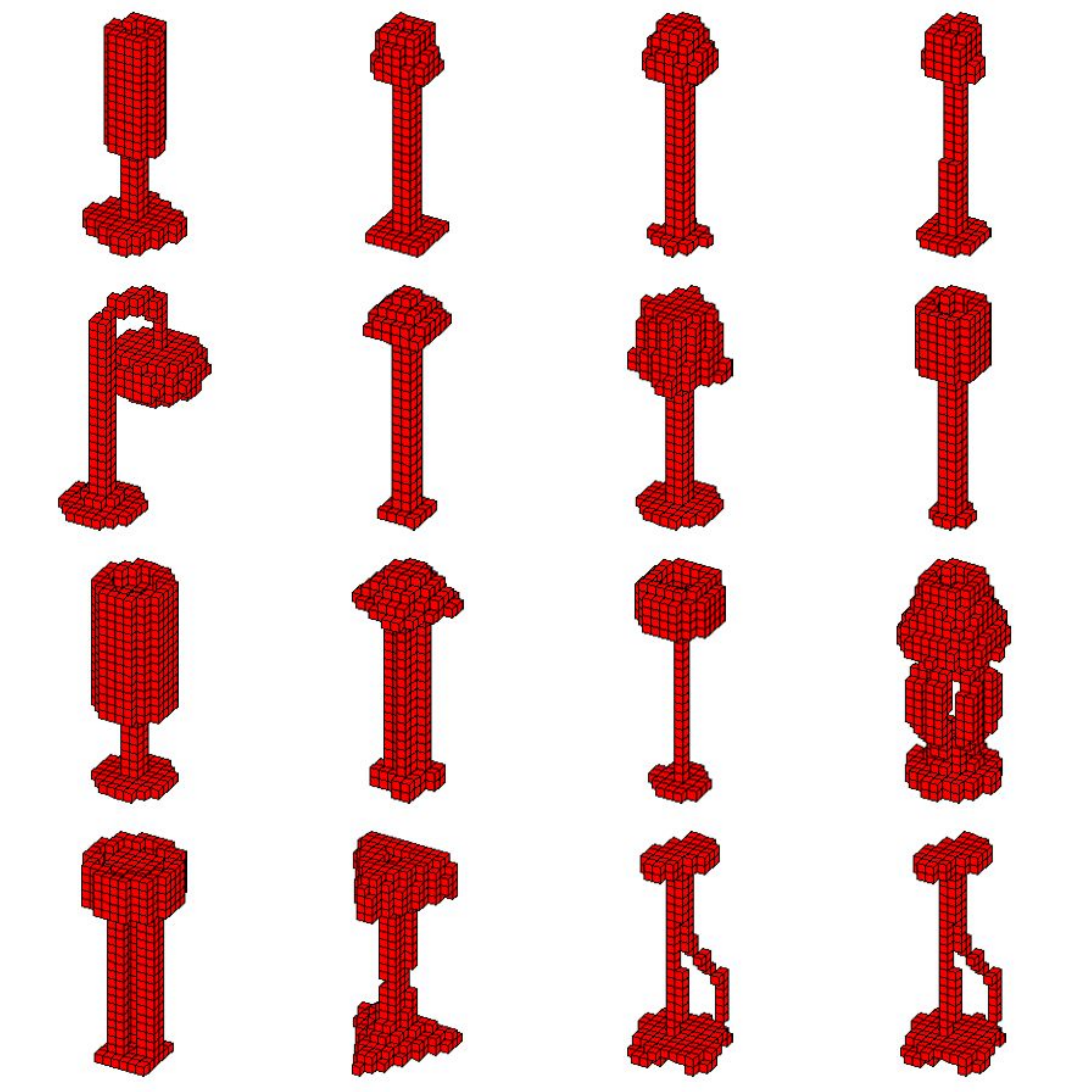}
\caption{ }
\end{subfigure}
\centering
\begin{subfigure}[b] {0.3\linewidth}
\centering
\includegraphics[width=0.95\linewidth]{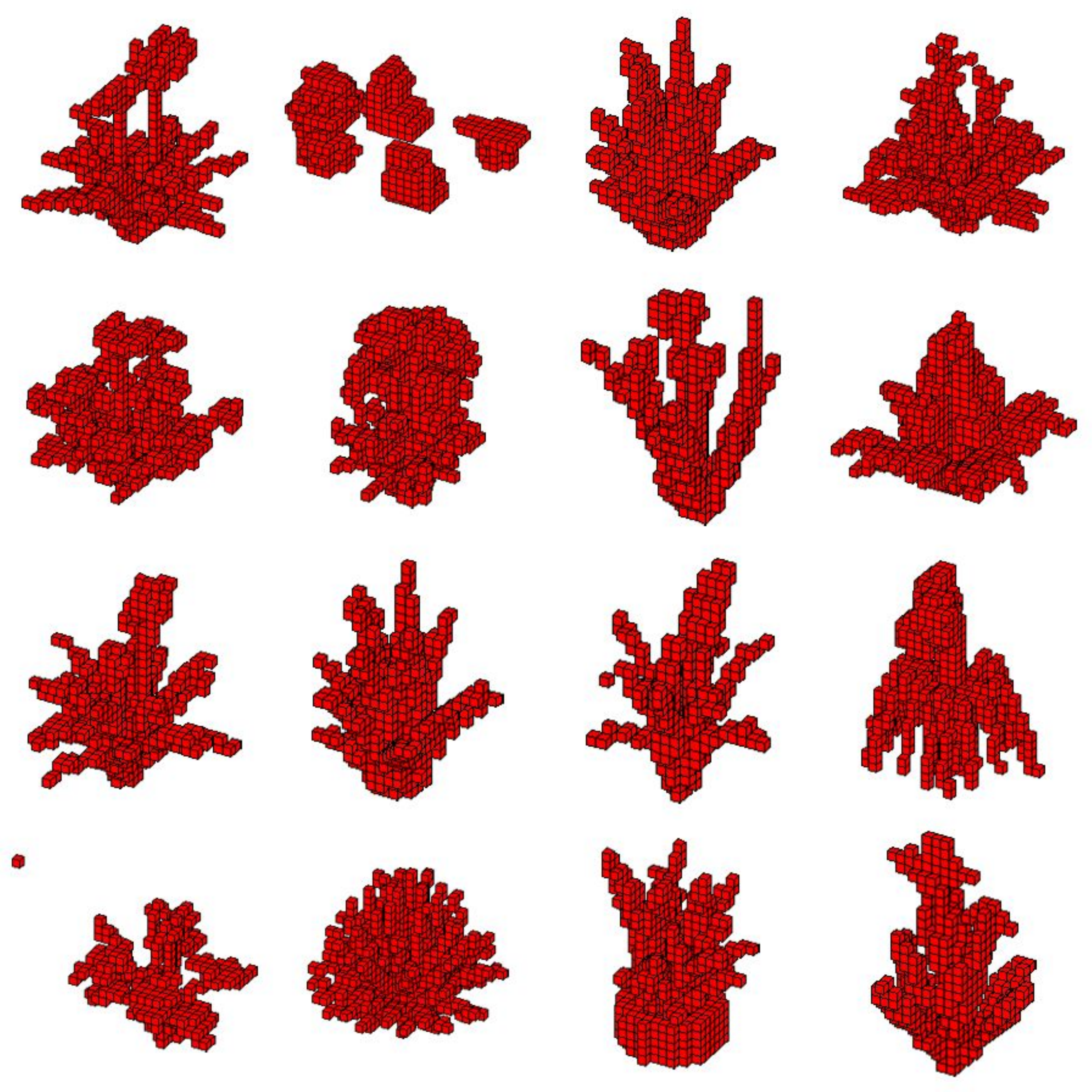}
\caption{ }
\end{subfigure}
\centering
\begin{subfigure}[b] {0.3\linewidth}
\centering
\includegraphics[width=0.95\linewidth]{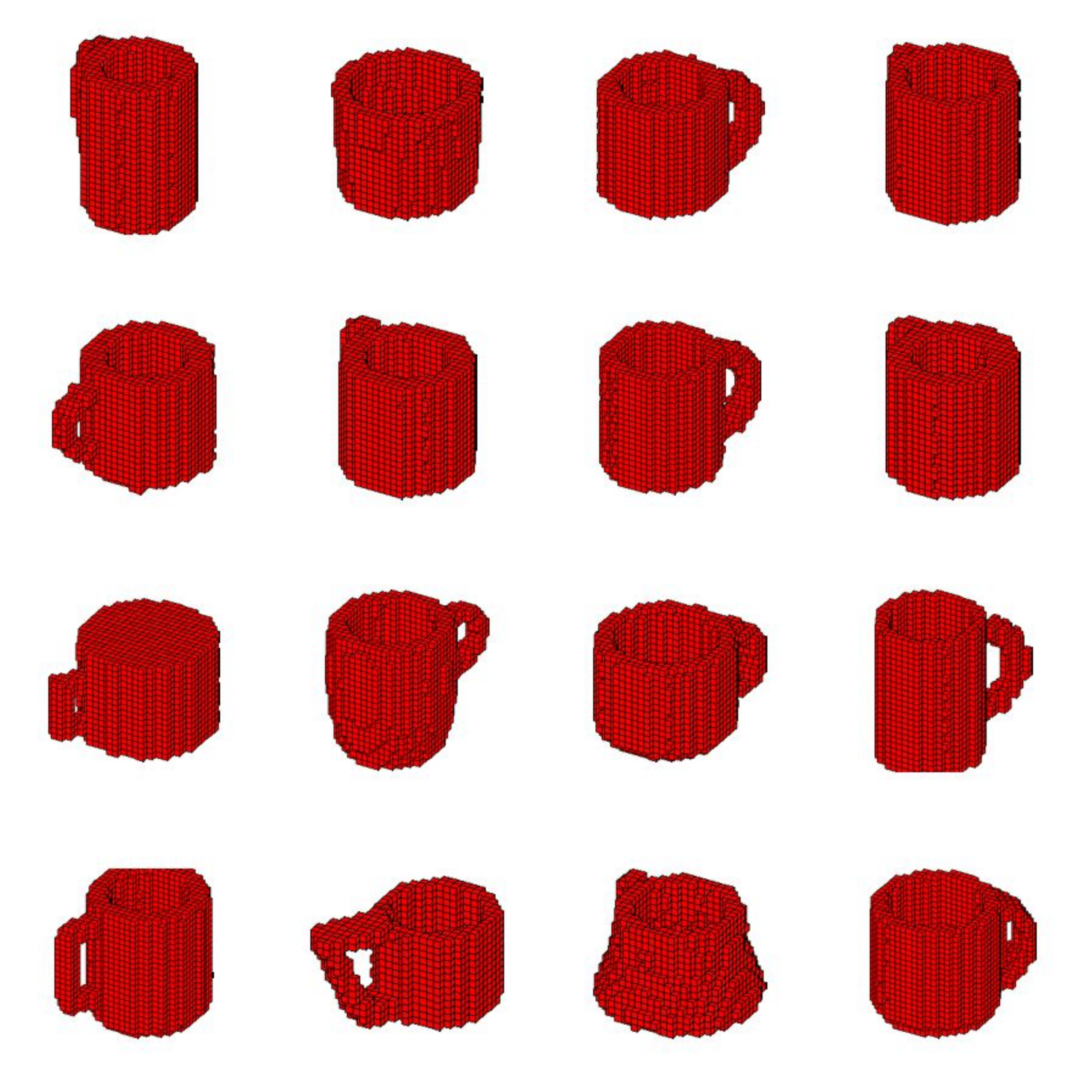}
\caption{ }
\end{subfigure}
\centering
\begin{subfigure}[b] {0.3\linewidth}
\centering
\includegraphics[width=0.95\linewidth]{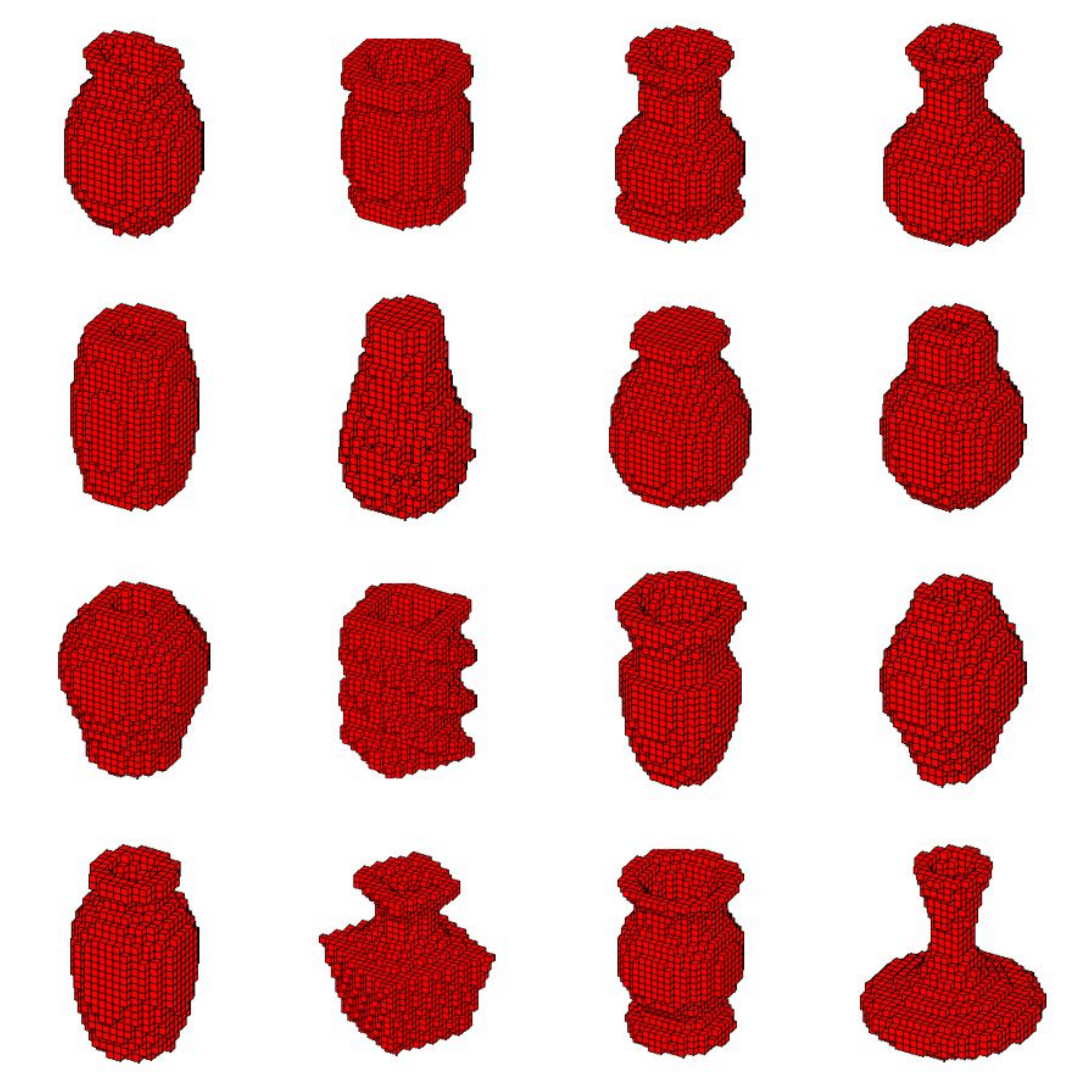}
\caption{ }
\end{subfigure}
\centering
\begin{subfigure}[b] {0.3\linewidth}
\centering
\includegraphics[width=0.95\linewidth]{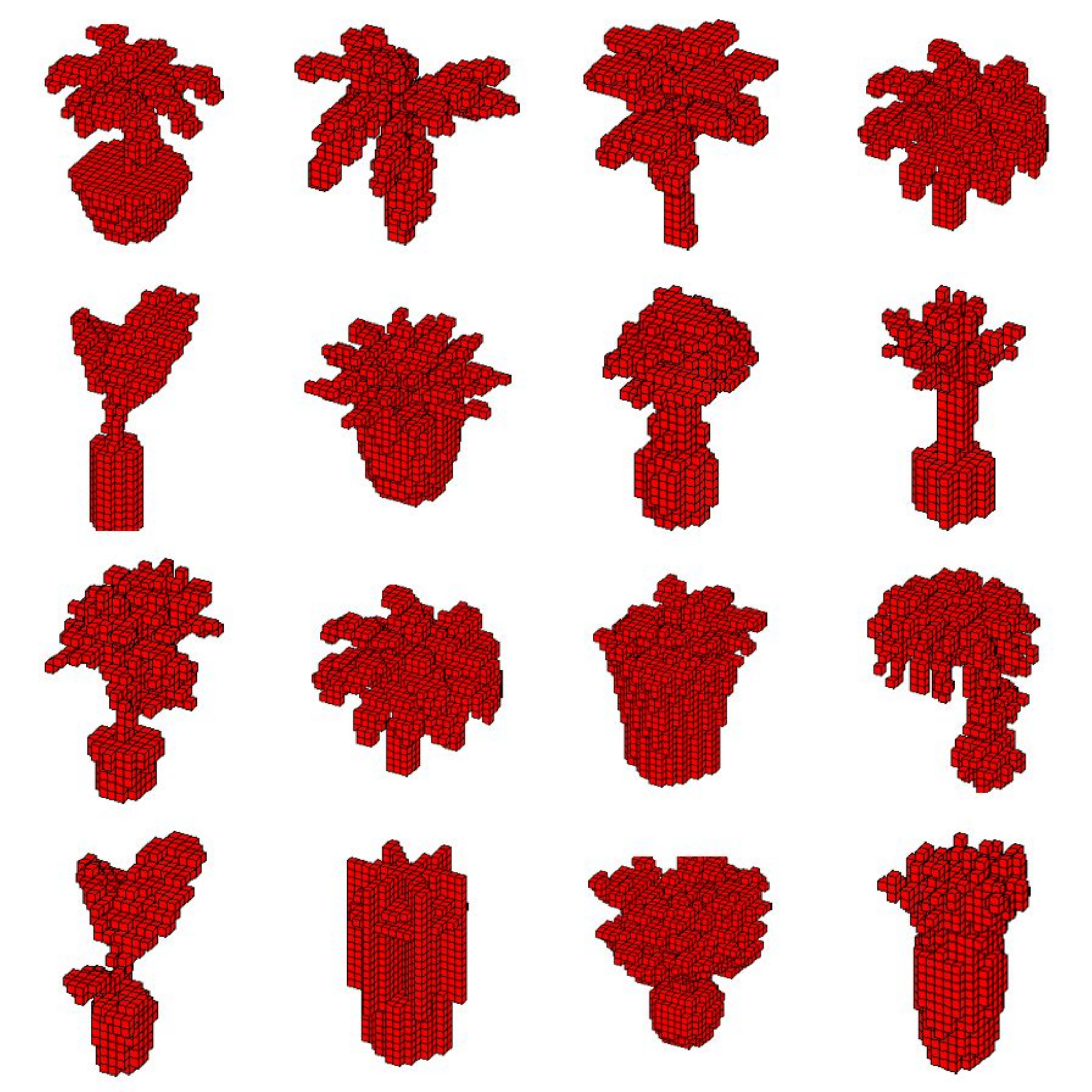}
\caption{ }
\end{subfigure}
\centering
\begin{subfigure}[b] {0.3\linewidth}
\centering
\includegraphics[width=0.95\textwidth]{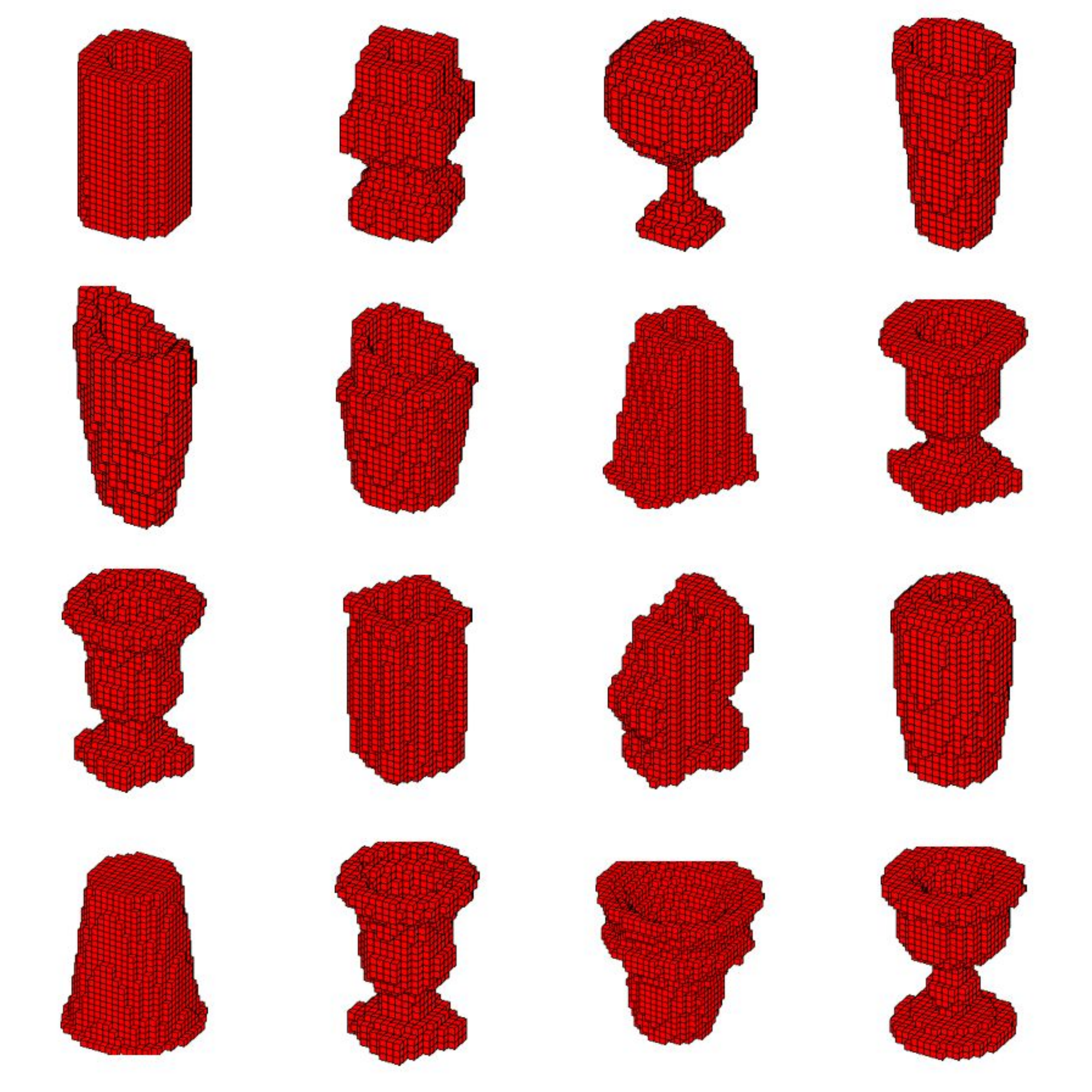}
\caption{ }
\end{subfigure}
\caption{The split of the confusion set of lamp, flower pot, cup and
vase yields three pure subsets (a) lamp, (b) plant and (c) cup and three
mixed subsets (d) flower pot and vase, (e) flower pot and plant and (f)
cup, flower pot, lamp, vase.}\label{fig.confClustering2}
\end{figure}
%%%%%%%%%%%%%%%%%%%%%%%%%%%%%%%%%%%%%%%%%%%%%%%%%%%%

{\bf Confusion Set Re-Classification.} We split a confusion set into
multiple subsets using the tree-structured hierarchical clustering
algorithm. Some of them contain samples from the same class while others
still contain samples from multiple classes. They form pure subsets and mixed subsets, respectively. We show splitting results
of two confusion sets in Fig.  \ref{fig.confClustering1} and Fig.
\ref{fig.confClustering2}, respectively. We see from the two figures
that shapes in pure subsets are distinctive from other subsets while
shapes in the mixed subsets share similar appearances. 

To deal with the challenging cases in each mixed subset, we need to
train a more effective classifier than the softmax classifier.
Furthermore, we have to avoid the potential overfitting problem due to a
very limited number of training samples.  For the above two reasons, we
choose the random forest classifier \cite{breiman2001random}. Classes
are weighted to overcome the unbalance sample distribution for each
mixed subset. In the testing phase, a sample is first assigned to a
subset based on its VCNN feature using the nearest neighbor rule. If it
is assigned to a pure subset, the class label is output as the predicted
label. Otherwise, we run the random forest classifier trained for the
particular mixed subset to make final class decision. 

\section{Experimental Results} \label{sec.experiment}

We test the network parameter selection algorithm and the confusion set
re-classification algorithm on the popular dataset ModelNet40
\cite{wu20153d}.  The ModelNet40 dataset contains 9,843 training samples
and 2,468 testing samples categorized into 40 classes. The original mesh
model is voxelized into dimension $30 \times 30 \times 30$ before
training and testing. The basic structure of the proposed VCNN is the
same as VoxNet \cite{maturana2015voxnet}: two convolutional layers
followed by one fully connected layer. However, we allow to have
different filter sizes and filter numbers at these layers.

\begin{figure}[t!]
\centering
\includegraphics[width=0.9\linewidth]{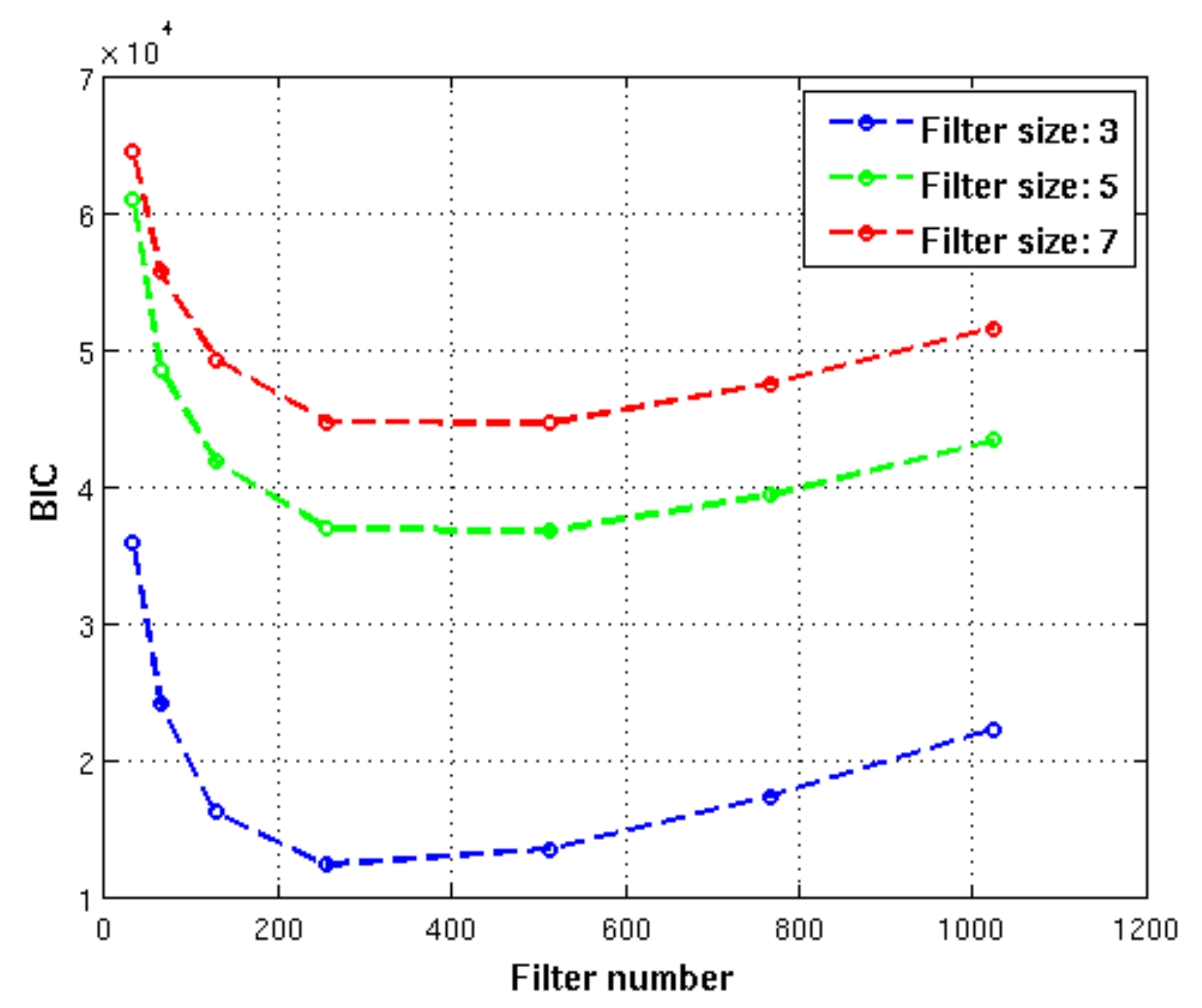}
\caption{Three BIC curves measured under three different filter sizes
$m_j = 3, 5, 7$ and seven different filter numbers $K_j = 32, 64, 128,
256, 512, 768, 1024$ for the first convolutional layer.}\label{fig.BICcurve}
\end{figure}

{\bf Results of Network Parameters Selection Alone.} We compare the BIC
scores with three spatial filter sizes $m^j = 3, 5, 7$ in the two
convolutional layer. The filter number is chosen to be $K^j =$ 32, 64,
128, 256, 512, 768, 1024. For the fully connected layer, the filter
number is set to $K^f = 128, 256, 512, 1024, 2048$. We show the BIC
scores by adjusting $m^j$ and $K^j$ at the first convolutional layer in
Fig.  \ref{fig.BICcurve}.  To keep the BIC scores at the same scale of
different filter sizes and training sample numbers, we normalize the BIC
function in Eq.  \ref{eq.BIC} by $(m^j)^3$ and keep $N$ independent of
$m^j$. Three valley points can be easily detected at $(m^1,K^1) = (3,
256)$, $(5,512)$ and $(7,512)$. This result is reasonable since we need
more filters to represent larger spatial patterns.  By comparing the
three curves, the choice of $(m^1,K^1) = (3,256)$ gives the lowest BIC
score. By adopting the greedy search algorithm introduced in Section
\ref{sec.design}, the network parameters in all layers of the proposed
VCNN are summarized in Table \ref{table.VCNNPara}. We also include the
network parameters of the VoxNet in the table for comparison. 

%%%%%%%%%%%%%%%%%%%%%%%%%%%%%%%%%%%%%%%%%%%%%%%%%%%%
%%%%%%%%%%%%%%%%%%%%%%%%%%%%%%%%%%%%%%%%%%%%%%%%%%%%

%%%%%%%%%%%%%%%%%%%%%%%%%%%%%%%%%%%%%%%%%%%%%%%%%%%
\begin{table}[t!]
\normalsize
\centering
\caption{Comparison of network parameters of the proposed VCNN and the
VoxNet, where the numbers in each cell follow the format $(m^j)^3 \times 
K^{j-1}.$}\label{table.VCNNPara}
%\resizebox{\linewidth}{!}{
\begin{tabular}{|c|c|c|c|c|}\hline
         & Conv1     & Conv2     & FC       & Output \\ \hline
VoxNet & $5^3\times 1 $  & $3^3 \times 32 $  & $6^3 \times 32$  & $1\times 1 \times 128$ \\ \hline
Our VCNN & $3^3\times 1 $  & $3^3 \times 256$  & $6^3 \times 128$ & $1\times 1 \times 1024$  \\ \hline
\end{tabular}%}
\end{table}
%%%%%%%%%%%%%%%%%%%%%%%%%%%%%%%%%%%%%%%%%%%%%%%%%%%%

%%%%%%%%%%%%%%%%%%%%%%%%%%%%%%%%%%%%%%%%%%%%%%%%%%%%
\begin{table}[t!]
\normalsize
\centering
\caption{Comparison of ACA and AIA scores of several state-of-the-art
methods for the ModelNet40 dataset. Our results are set to bold. }\label{table.ACAAIA}
%\resizebox{0.85\linewidth}{!}{
\begin{tabular}{|c|c|c|}
\hline
Volume-based methods                 & ACA              & AIA              \\ \hline
3DShapeNets\cite{wu20153d}            & 77.30\%          & -                \\ \hline
VoxNet\cite{maturana2015voxnet}               & 83.01\%          & 87.40\%          \\ \hline
3D-Gan \cite{wu2016learning}                & 83.30\%          & -                \\ \hline
SubVolume\cite{qi2016volumetric}              & 86.00\%          & 89.20\%          \\ \hline
AniProbing \cite{qi2016volumetric}            & 85.60\%          & 89.90\%          \\ \hline
\textbf{VCNN /wo ReC}       & \textbf{85.66\%} & \textbf{89.34\%} \\ \hline
\textbf{VCNN /w ReC} & \textbf{86.23\%} & \textbf{89.78\%} \\ \hline \hline
View-based methods                 & ACA              & AIA              \\ \hline 
DeepPano \cite{shi2015deeppano}              & 77.63\%          & -                \\ \hline
GIFT \cite{bai2016gift}                  & 83.10\%          & -                \\ \hline
MVCNN \cite{su2015multi}                 & 90.10\%          & -                \\ \hline
Pairwise \cite{johns2016pairwise}              & 90.70\%           & -                \\ \hline
FusionNet \cite{hegde2016fusionnet}             & 90.80\%           & -                \\ \hline
\end{tabular}%}
\end{table}
%%%%%%%%%%%%%%%%%%%%%%%%%%%%%%%%%%%%%%%%%%%%%%%%%%%%

%%%%%%%%%%%%%%%%%%%%%%%%%%%%%%%%%%%%%%%%%%%%%%%%%%%%
\begin{figure}[t!]
\centering
\includegraphics[width=\linewidth]{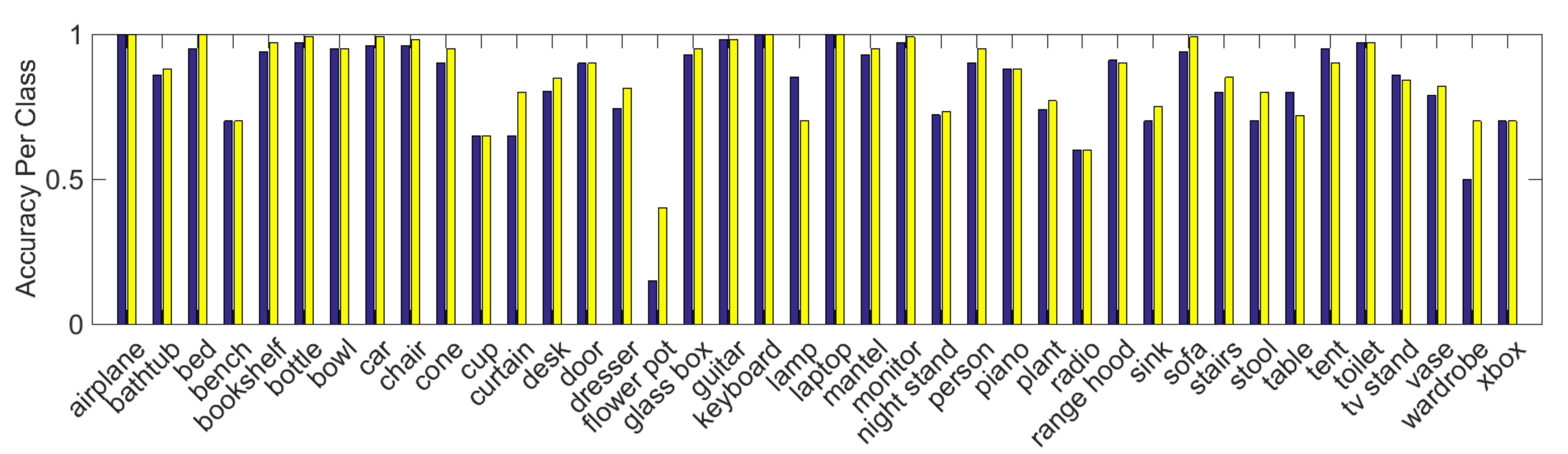}
\caption{Comparison of accuracy per classe between the VoxNet method
(blue bars) and our VCNN design (yellow bars).}\label{fig.AccPerClass}
\vspace{-1.5em}
\end{figure}
%%%%%%%%%%%%%%%%%%%%%%%%%%%%%%%%%%%%%%%%%%%%%%%%%%%%

To compare the performance of the proposed VCNN and other benchmarking
methods, we use two performance measures: average class accuracy (ACA)
and average instance accuracy (AIA). The ACA score averages the prediction accuracy scores over classes, while the AIA score takes the average of prediction scores over testing samples. The results are shown in Table
\ref{table.ACAAIA}, where methods are classified into the view-based and
the volume-based two categories.  For the proposed VCNN, we consider two
cases: with and without confusion set reclassification. We first focus
on the contribution of the network parameters selection algorithm
without the confusion set reclassification (w/o ReC) module.  The VCNN
w/o ReC outperforms the VoxNet by $2.65\%$ and $1.94\%$ in ACA and
AIA, respectively. The gap between volume-based and view-based methods
is narrowed. Although the ACA performance of the VCNN w/o ReC is lower
than that of the SubVolume method by $0.34\%$ and its AIA performance is
lower than that of the AniProbing method by $0.56\%$, these
two methods use more complex network structures such as
multi-orientation pooling and subvolume supervision. 

It is worthwhile to examine the classification accuracy for each
individual class. We compare the accuracy of the proposed VCNN with that
of the VoxNet in Fig.  \ref{fig.AccPerClass}. By leveraging the
network parameters selection, the performance of most classes is boosted
with VCNN.  Furthermore, classes of low classification accuracy belong
to some confusion sets. The observed results are consistent with those
shown in Fig. \ref{fig.confgroups}. 

%%%%%%%%%%%%%%%%%%%%%%%%%%%%%%%%%%%%%%%%%%%%%%%%%%%%
\begin{figure}[th!]
\centering
\begin{subfigure}[b] {0.4\linewidth}
\centering
\includegraphics[width=0.3\linewidth]{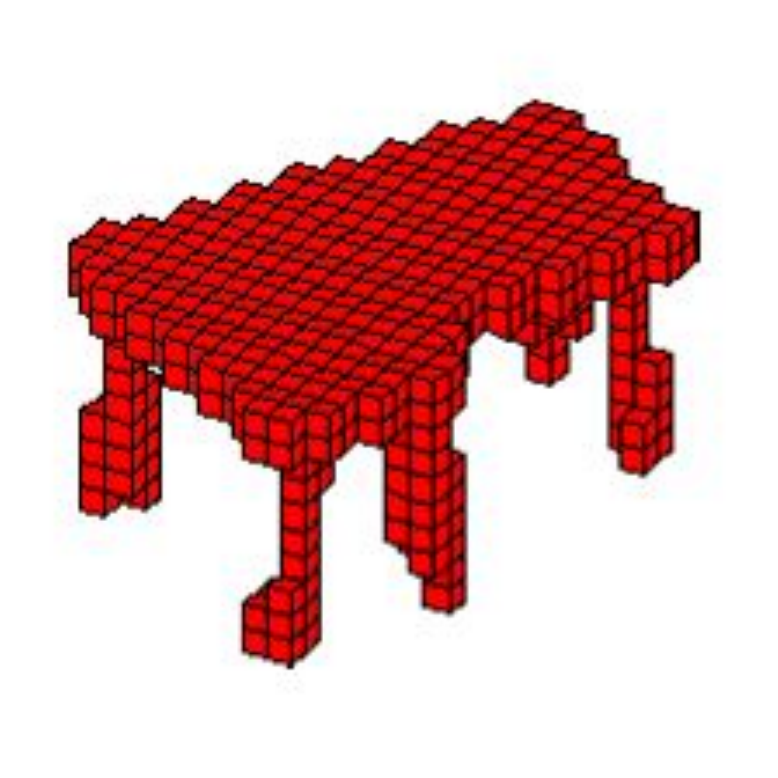}
\includegraphics[width=0.9\linewidth]{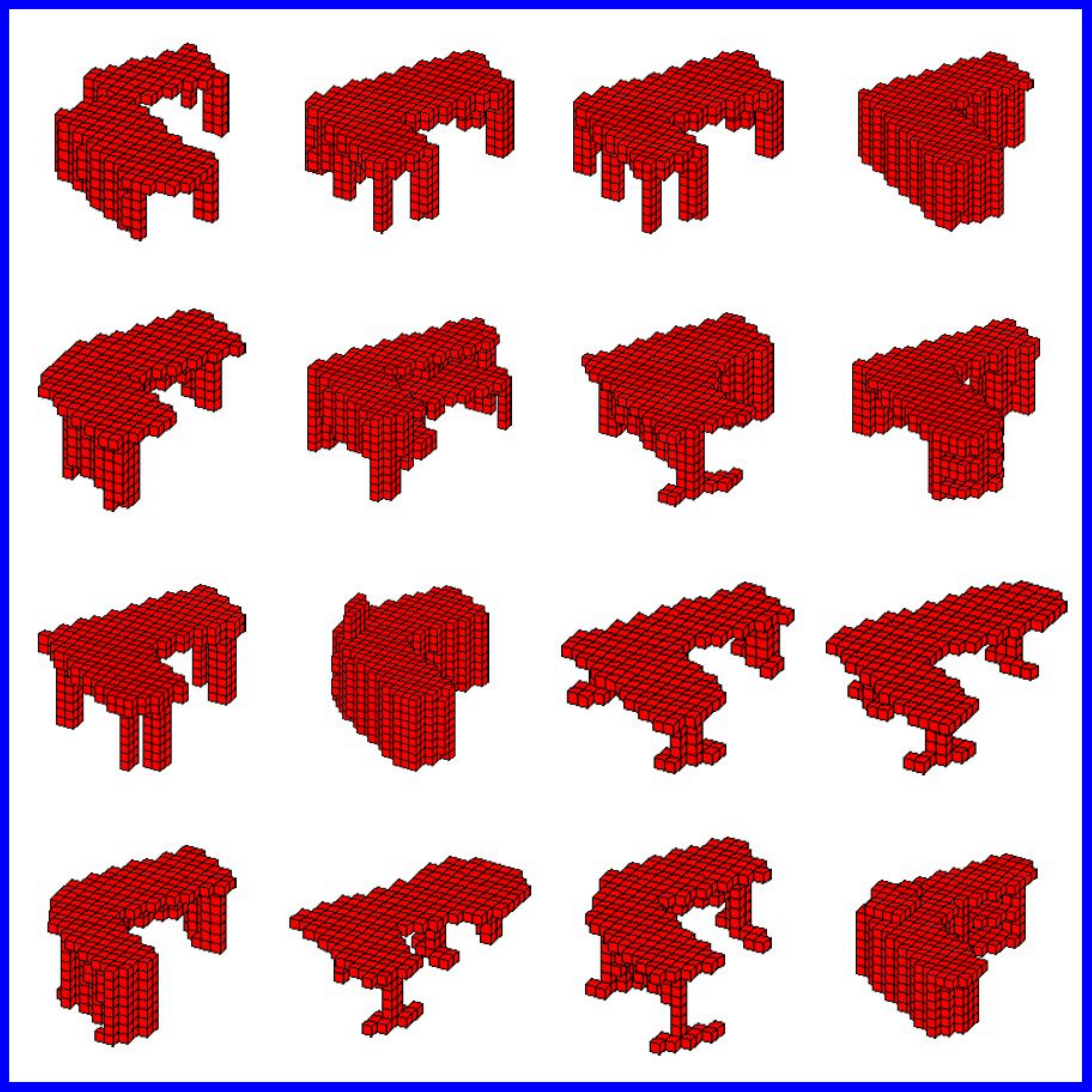}
\caption{Desk}
\end{subfigure}
\centering
\begin{subfigure}[b] {0.4\linewidth}
\centering
\includegraphics[width=0.3\linewidth]{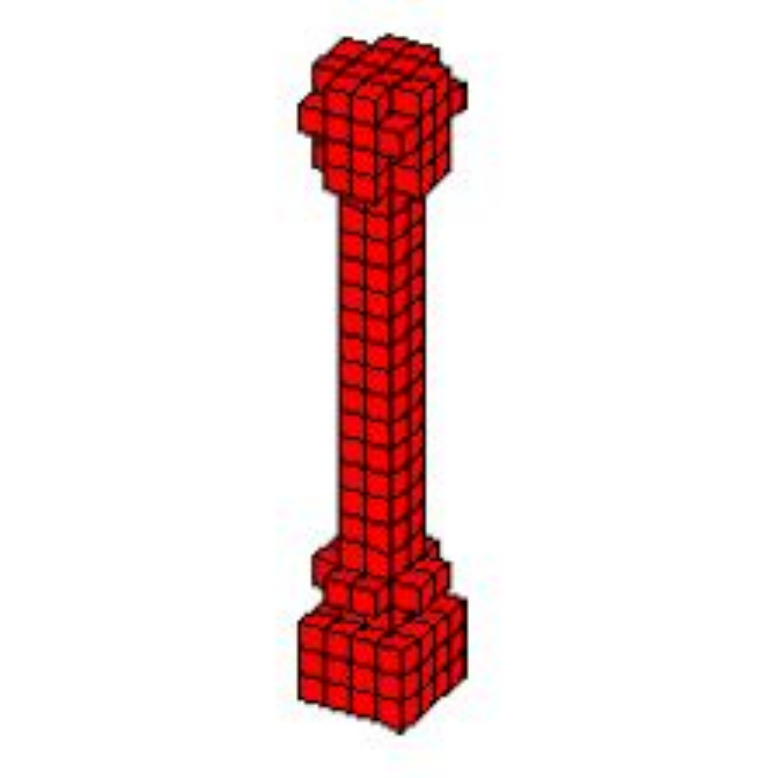}
\includegraphics[width=0.9\linewidth]{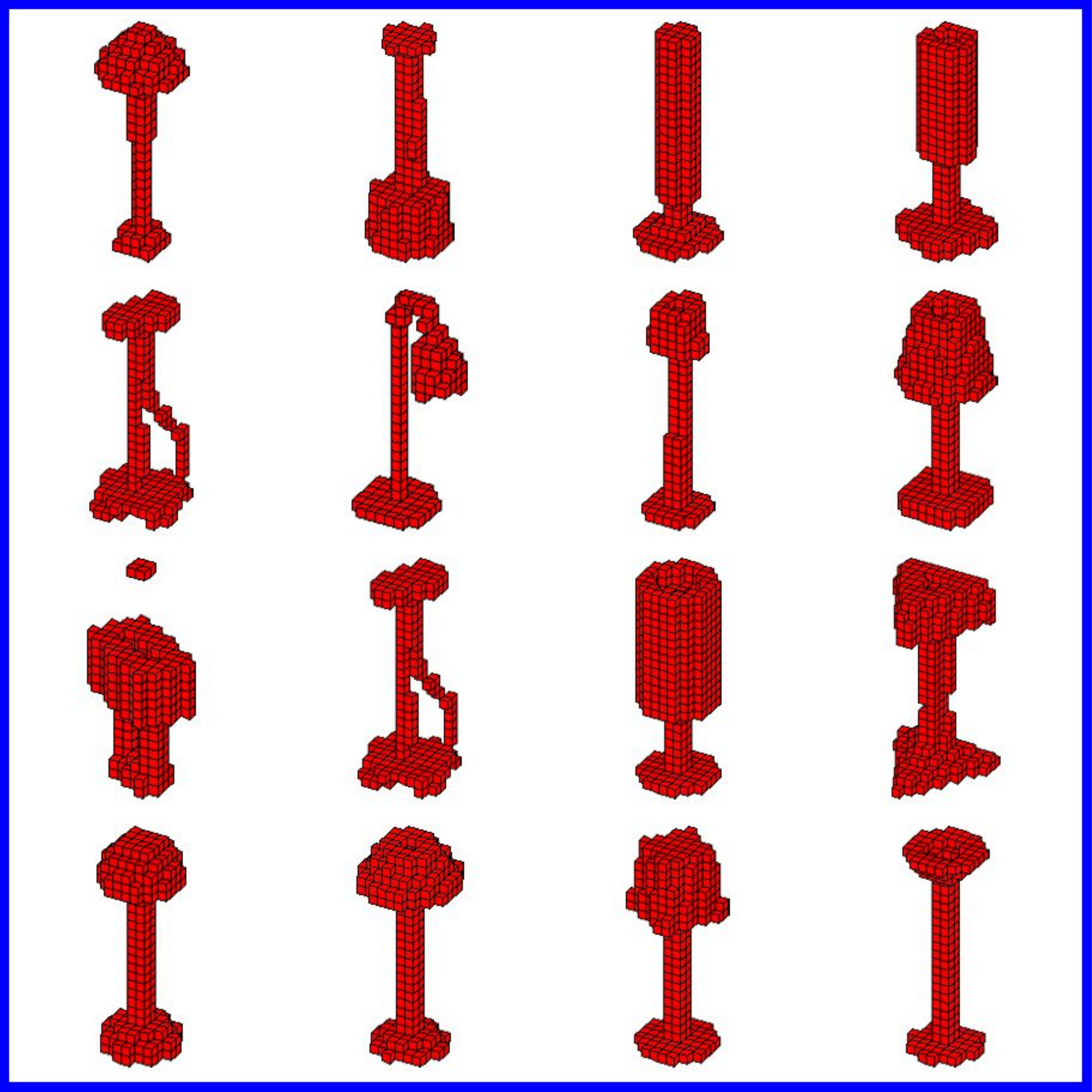}
\caption{Lamp}
\end{subfigure}
\centering
\begin{subfigure}[b] {0.4\linewidth}
\centering
\includegraphics[width=0.3\linewidth]{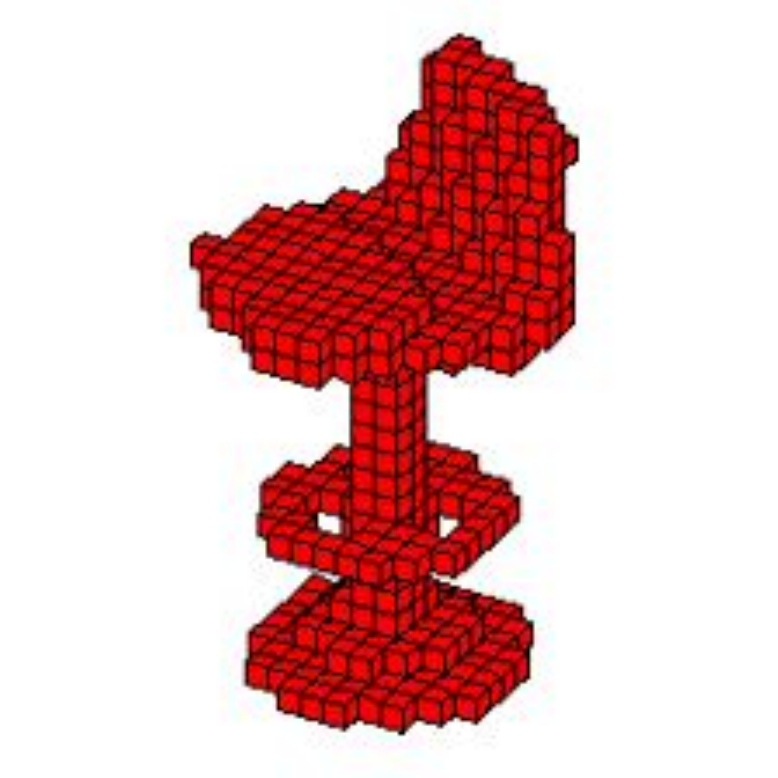}
\includegraphics[width=0.9\linewidth]{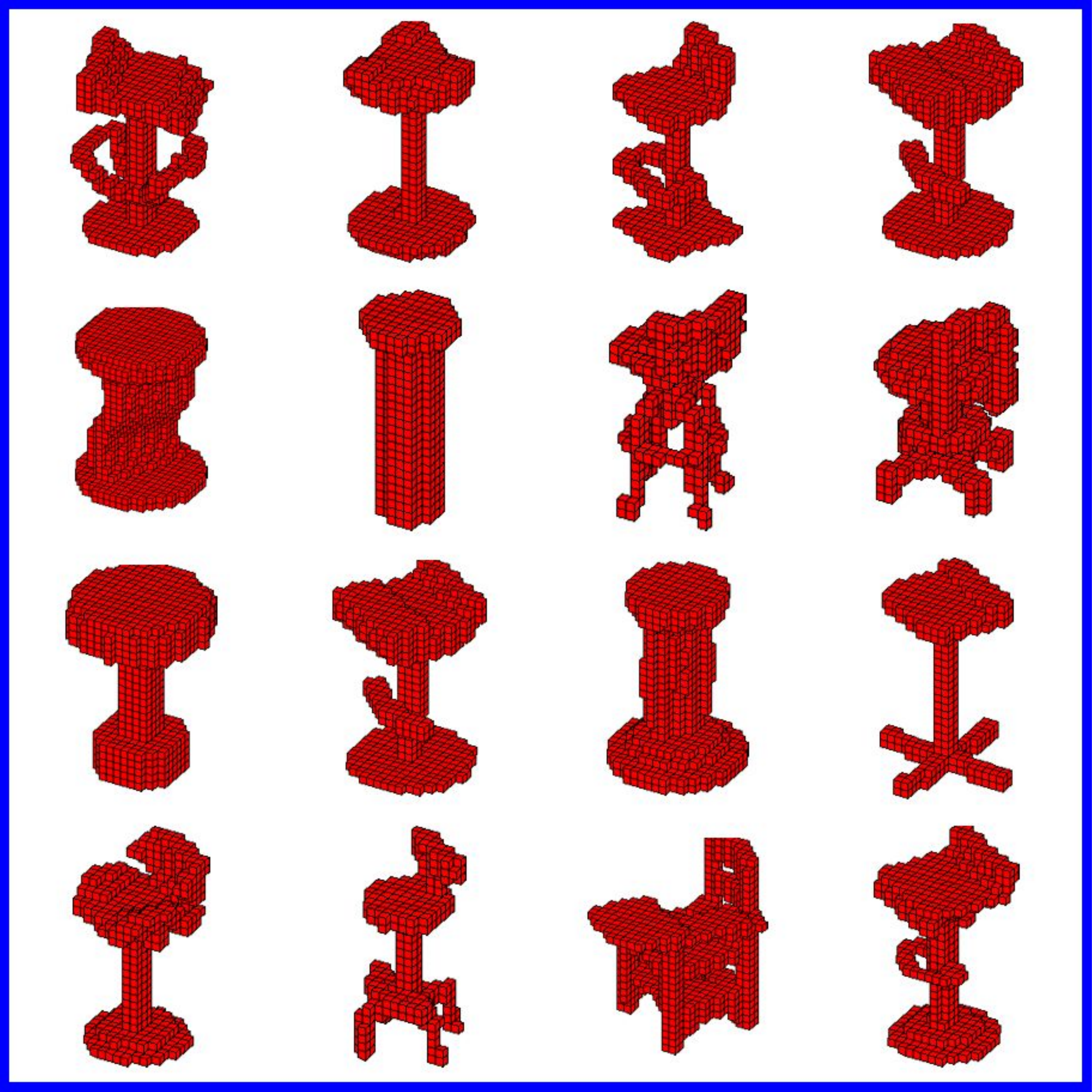}
\caption{Chair}
\end{subfigure}
\centering
\begin{subfigure}[b] {0.4\linewidth}
\centering
\includegraphics[width=0.3\linewidth]{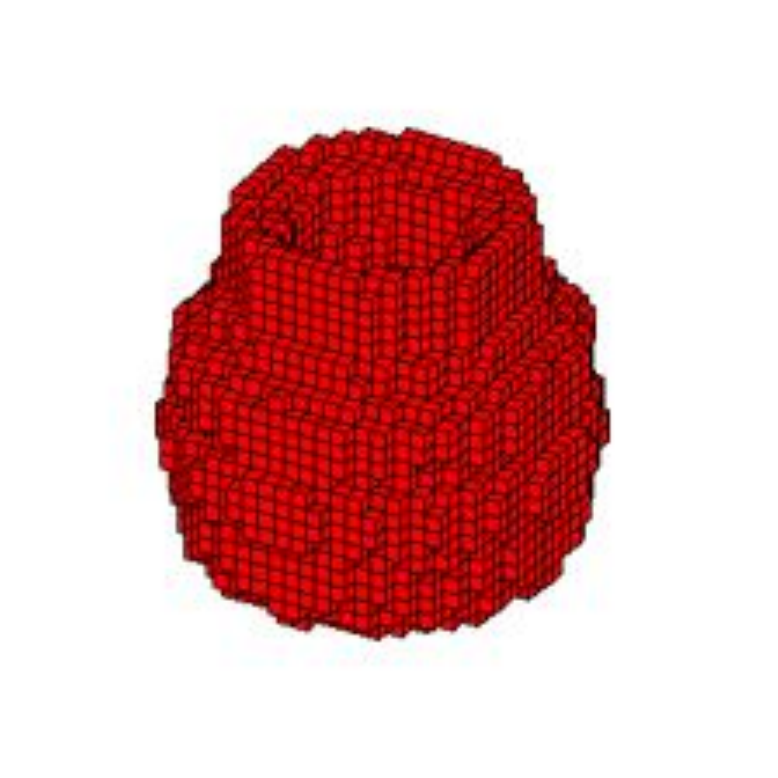}
\includegraphics[width=0.9\linewidth]{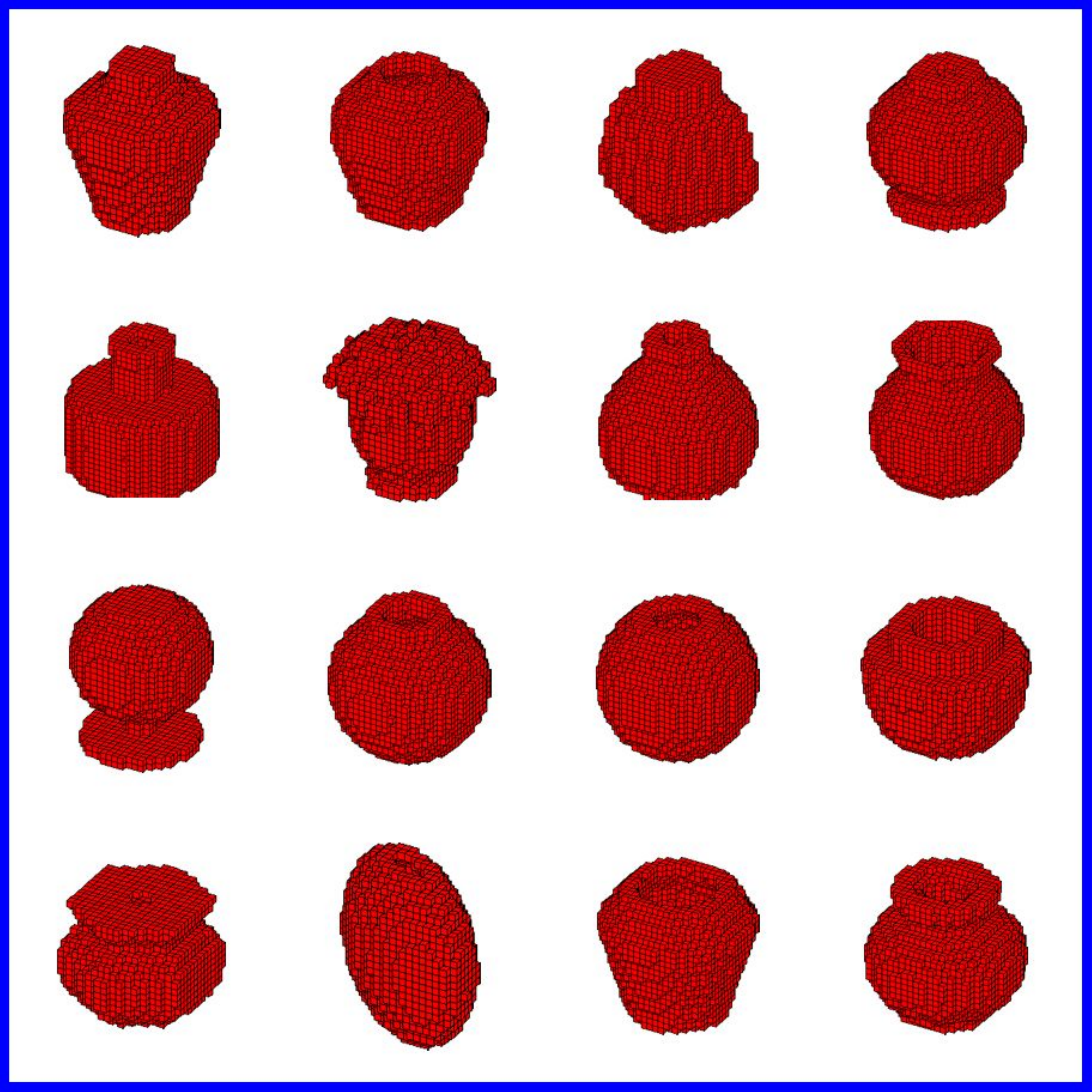}
\caption{Vase}
\end{subfigure}
\caption{Four examples of corrected errors: (a) desk, (b) lamp, (c)
chair, (d) vase. Each example has a testing sample on the top and
its assigned subset on the bottom. }\label{fig.wrong2correct}
\vspace{-1.5em}
\end{figure}
%%%%%%%%%%%%%%%%%%%%%%%%%%%%%%%%%%%%%%%%%%%%%%%%%%%%

%%%%%%%%%%%%%%%%%%%%%%%%%%%%%%%%%%%%%%%%%%%%%%%%%%%%
\begin{figure}[ht!]
\centering
\begin{subfigure}[b] {0.4\linewidth}
\centering
\includegraphics[width=0.3\linewidth]{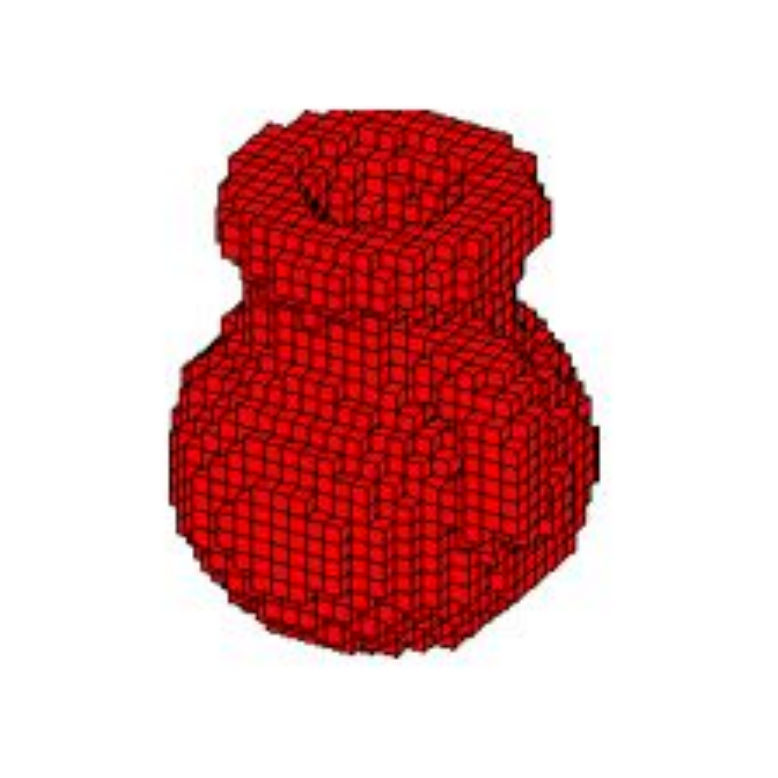}
\includegraphics[width=0.9\linewidth]{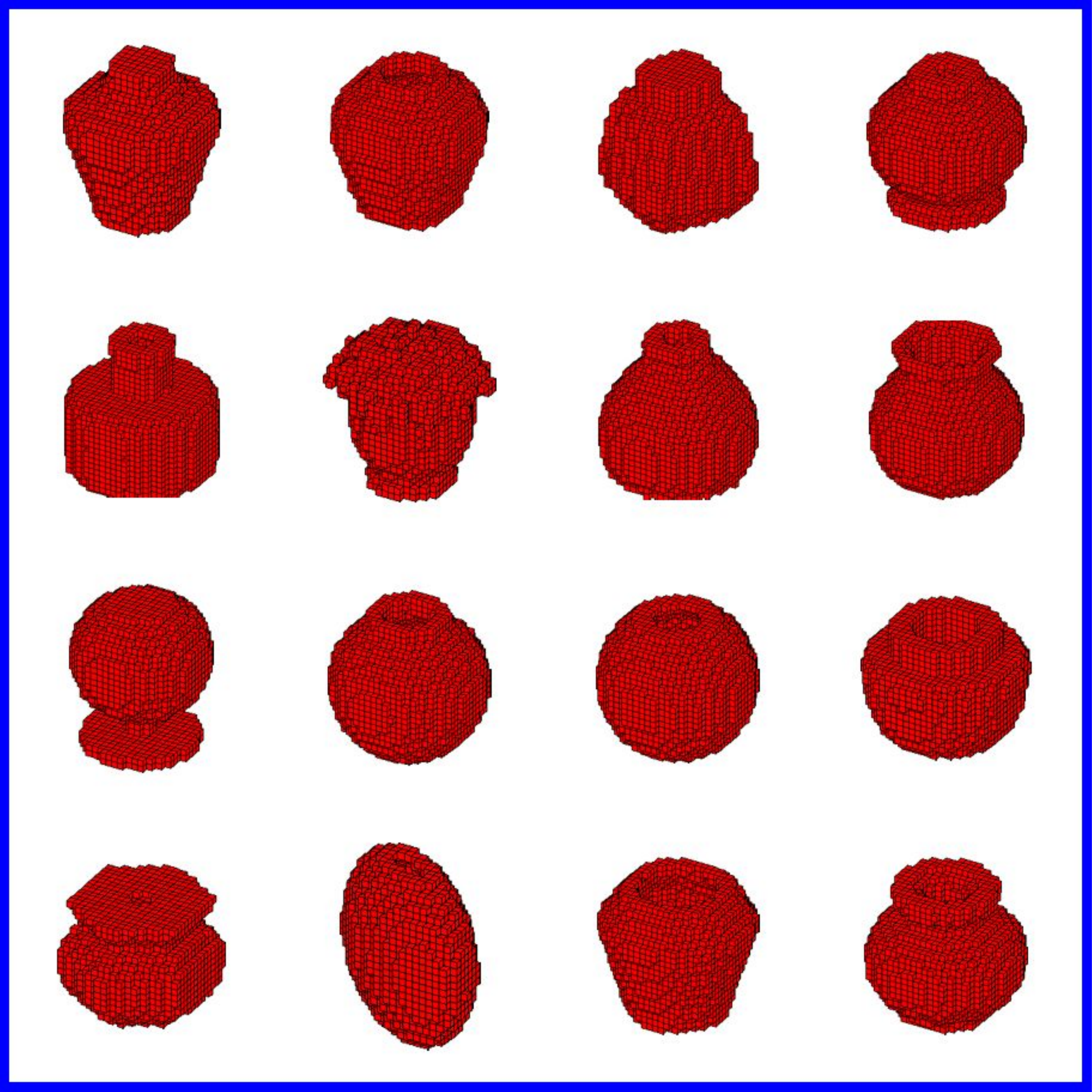}
\caption{Cup}
\end{subfigure}
\centering
\begin{subfigure}[b] {0.4\linewidth}
\centering
\includegraphics[width=0.3\linewidth]{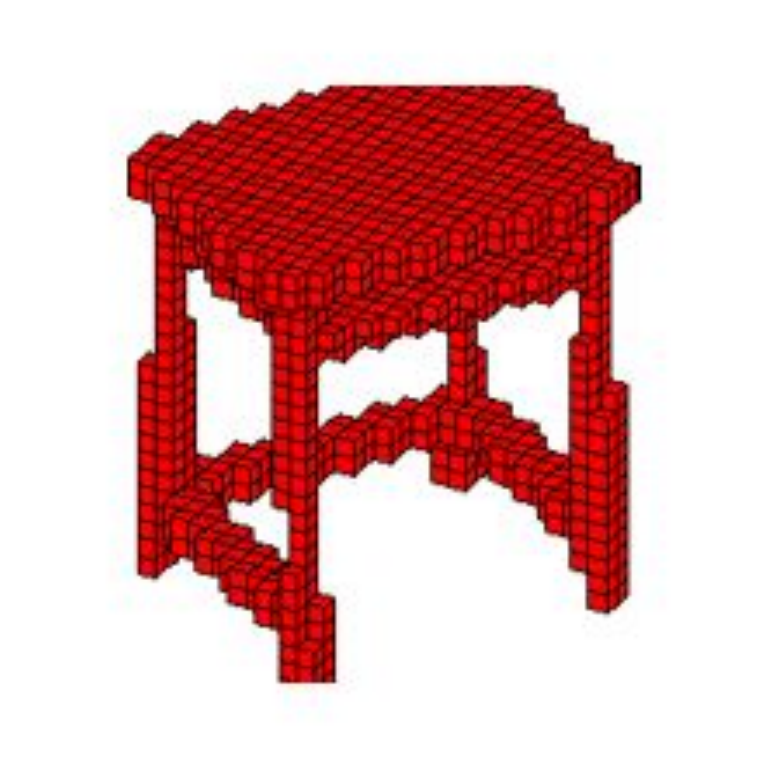}
\includegraphics[width=0.9\linewidth]{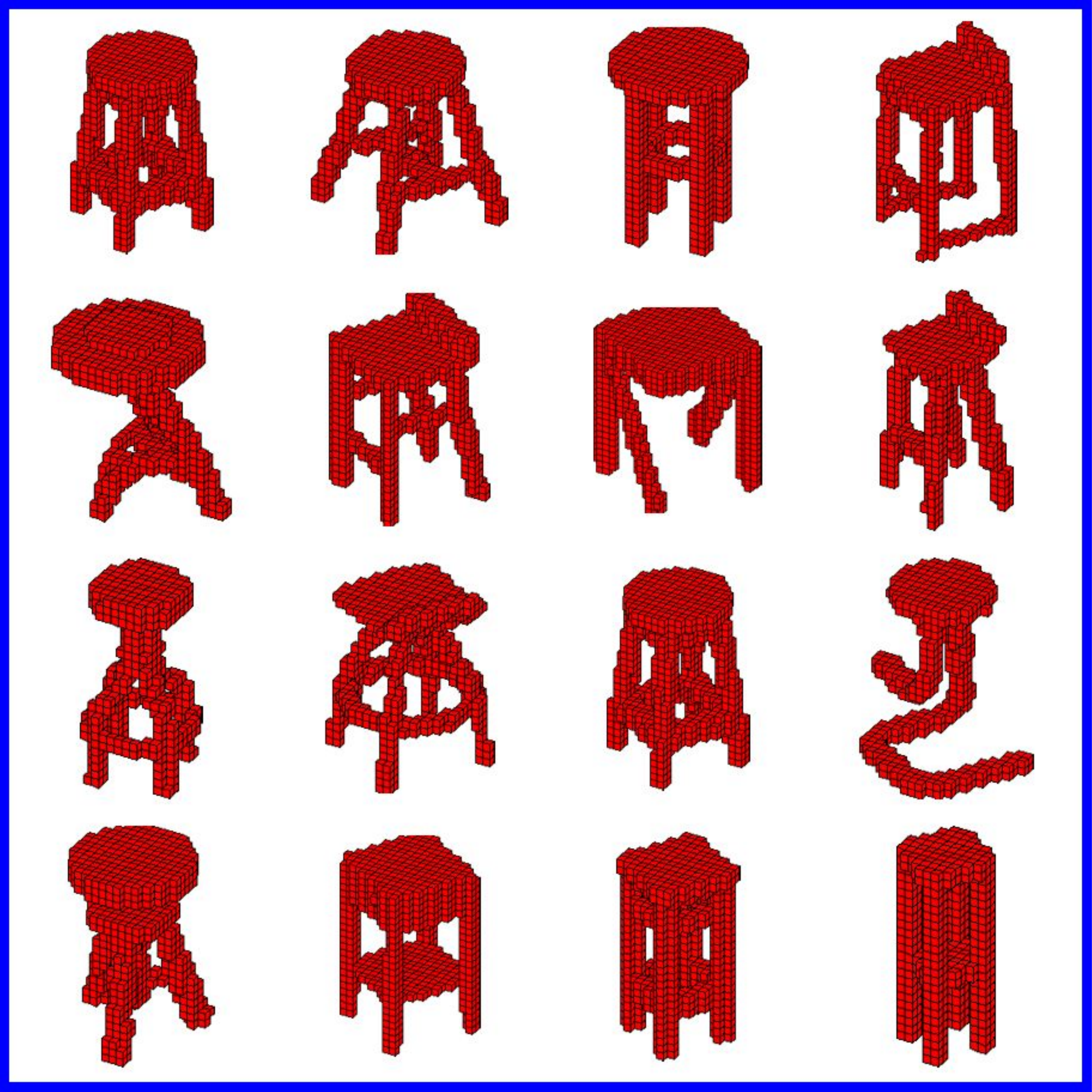}
\caption{Desk}
\end{subfigure}
\centering
\begin{subfigure}[b] {0.4\linewidth}
\centering
\includegraphics[width=0.3\linewidth]{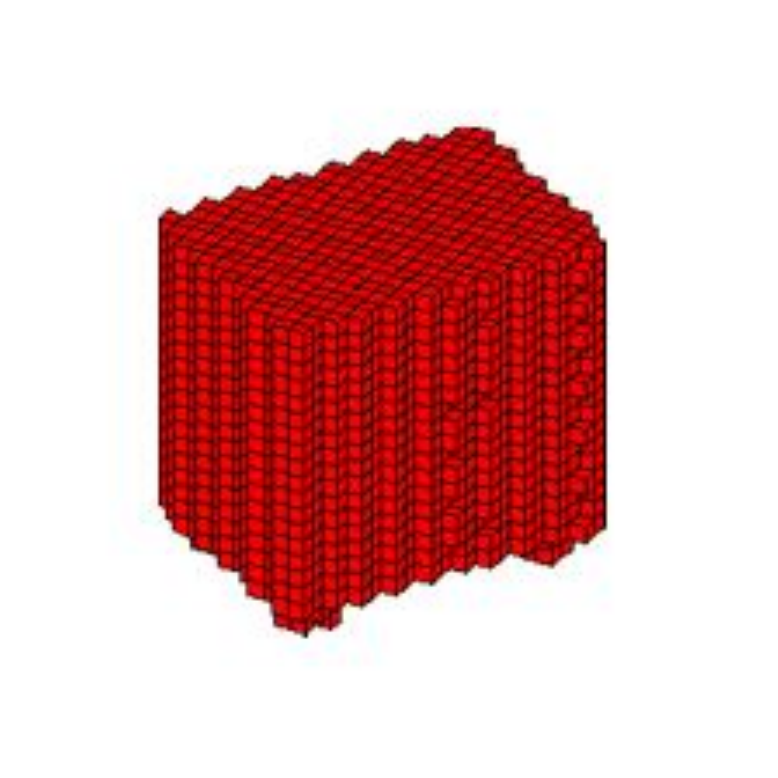}
\includegraphics[width=0.9\linewidth]{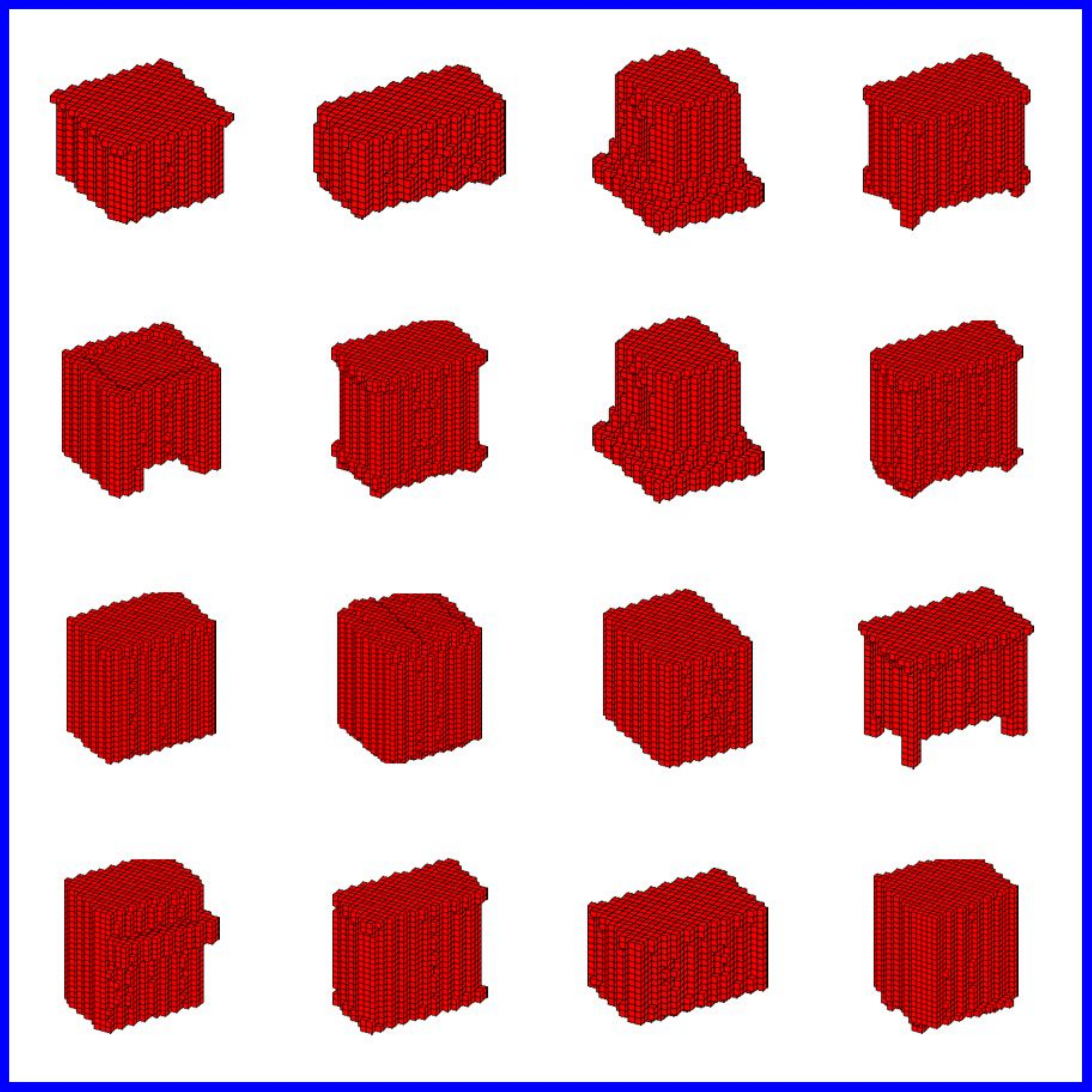}
\caption{Dresser}
\end{subfigure}
\centering
\begin{subfigure}[b] {0.4\linewidth}
\centering
\includegraphics[width=0.3\linewidth]{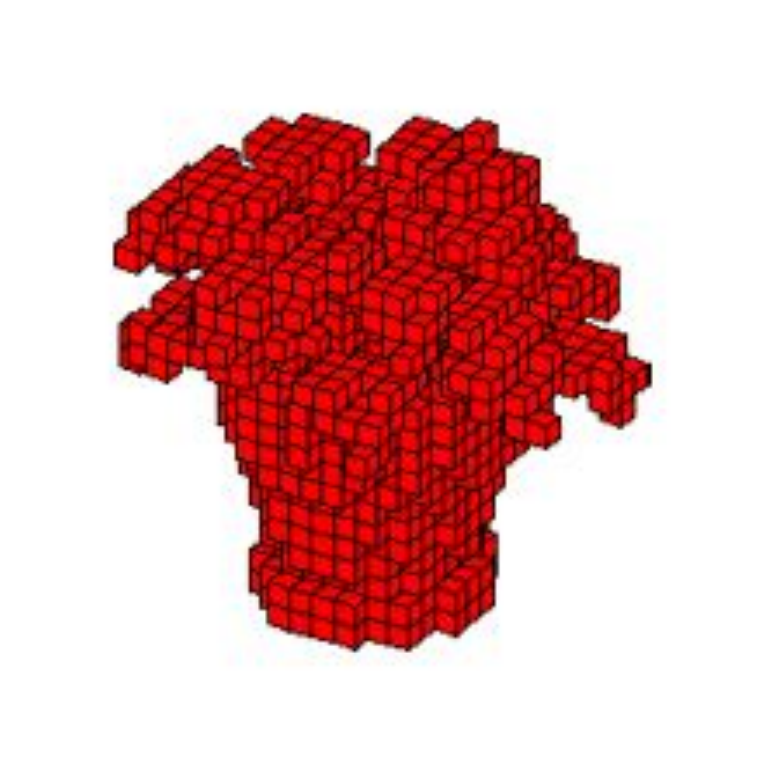}
\includegraphics[width=0.9\linewidth]{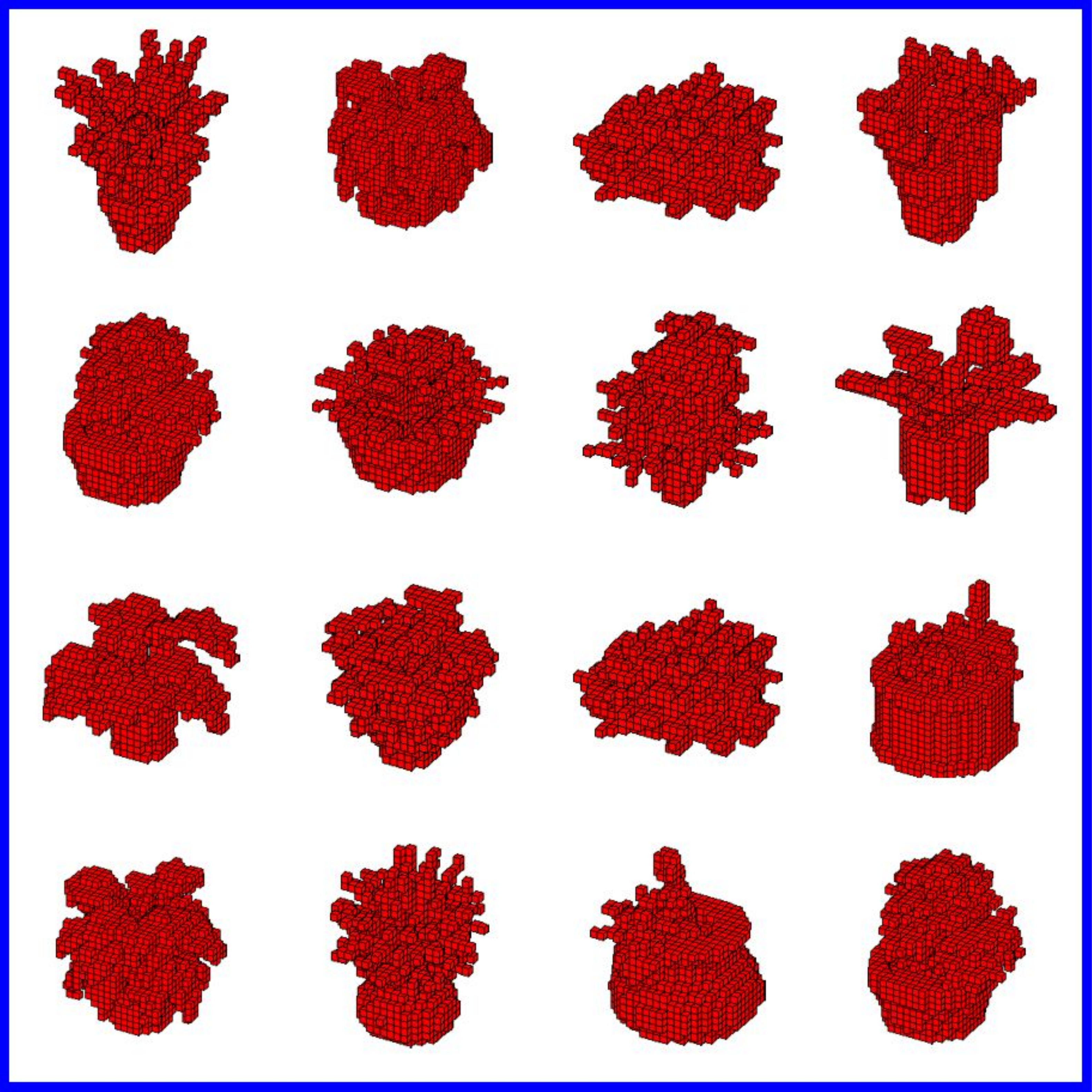}
\caption{Plant}
\end{subfigure}
\caption{Four examples of uncorrected errors: (a) cup, (b) desk, (c)
dresser, (d) plant. Each example includes a testing sample on the top
and the assigned subset on the bottom. }\label{fig.wrong2wrong}
\vspace{-1.5em}
\end{figure}
%%%%%%%%%%%%%%%%%%%%%%%%%%%%%%%%%%%%%%%%%%%%%%%%%%%%

{\bf Results of Confusion Set Re-Classification.} As shown in Table
\ref{table.ACAAIA}, the classification accuracy can be boosted
furthermore by the proposed re-classification algorithm. The improved
ACA and AIA values are $0.57\%$ and $0.44\%$.  Since the error cases in
mixed subsets are difficult cases, such performance gains are good. We
observe two cases in the improvement, where wrongly predicted samples
can be corrected. First, a sample is assigned to a correct pure subset
so that its prediction can be corrected directly.  Examples are given in
Figs.  \ref{fig.wrong2correct} (a) and (b), where a desk shape and a
lamp shape are correctly assigned to one of their pure subsets,
respectively.  Second, a sample is assigned to a mixed subset and the
random forest classifier helps correct the prediction result. Examples
are given in Figs. \ref{fig.wrong2correct} (c) and (d).  A chair shape
is assigned to a mixed subset containing chairs and stools in Fig.
\ref{fig.wrong2correct} (c). A vase shape is assigned a mixed subset
containing flower pots and vases.  in Fig.  \ref{fig.wrong2correct} (d).
They can be eventually correctly classified using the random forest
classifier.  The power of the confusion set identification and
re-classification procedure is demonstrated by these examples. 

Although errors still exist, they can be clearly analyzed based on our
clustering results. Uncorrected errors come from the strong feature
similarity. In Fig. \ref{fig.wrong2wrong} (a) and (b), a cup shape and a
desk shape are wrongly assigned to a mixed subset of vase and flower pot
and a mixed subset of chair and stool, respectively. In Fig.
\ref{fig.wrong2wrong}(c), although a dresser shape is assigned to a
mixed subset containing dressers, night stands, radio, tv stands and
wardrobe, the VCNN features are not distinctive enough to produce a
correct classification. Similarly in Fig.  \ref{fig.wrong2wrong} (d), a
plant shape cannot be differentiated from the flower pot in its mixed
subset. These mistakes are due to their highly visual similarity with
shapes in other classes. Additionally, the success of our confusion set
identification algorithm offers the possibility of using more advanced
methods (e.g., a part-based classifier \cite{zhang2014part}) to improve
the classification performance on erroneous samples. A higher
classification accuracy is expected in the future. 

\section{Conclusion} \label{sec.conclusion}

The design, analysis and application of a volumetric convolutional
neural network (VCNN) were presented.  We proposed a feed-forward
K-means clustering algorithm to determine the filter number and size.
The cause of confusion sets was identified.  Furthermore, a hierarchical
clustering method followed by a random forest classification method was
proposed to boost the classification performance among confusing
classes. Finally, experiments were conducted on a popular ModelNet40
dataset. The proposed VCNN offers the state-of-the-art performance among
all volume-based CNN methods. 

\section*{Acknowledgment}

Computation for the work described in this paper was supported by the
University of Southern California's Center for High-Performance
Computing (hpc.usc.edu).

%% The Appendices part is started with the command \appendix;
%% appendix sections are then done as normal sections
%% \appendix

%% \section{}
%% \label{}

%% If you have bibdatabase file and want bibtex to generate the
%% bibitems, please use
%%
%\section{\refname}
%\section*{\refname}
\bibliographystyle{elsarticle-num} 
\bibliography{mybib}

%% else use the following coding to input the bibitems directly in the
%% TeX file.

%\begin{thebibliography}{00}

%% \bibitem[Author(year)]{label}
%% Text of bibliographic item

%\bibitem[ ()]{}

%\end{thebibliography}
\end{document}